\documentclass[runningheads]{llncs}

\usepackage{eccv}

\usepackage{eccvabbrv}

\usepackage{graphicx}
\usepackage{booktabs}
\usepackage{enumitem}
\usepackage{wrapfig}

\usepackage[dvipsnames]{xcolor}
\newcommand{\red}[1]{{\color{red}#1}}
\newcommand{\blue}[1]{{\color{blue}#1}}
\usepackage[accsupp]{axessibility}  

\usepackage[pagebackref, breaklinks,colorlinks,citecolor=eccvblue]{hyperref}

\usepackage{orcidlink}

\begin{document}

\title{RoCOCO: Robustness Benchmark of MS-COCO \\
to Stress-test Image-Text Matching Models} 

\titlerunning{RoCOCO: Robustness Benchmark of MS-COCO}

\author{Seulki Park\inst{1} \quad
Daeho Um\inst{2} \quad
Hajung Yoon\inst{2} \\
Sanghyuk Chun\inst{3} \quad
Sangdoo Yun\inst{3} \quad
Jin Young Choi\inst{2}
}

\authorrunning{S.~Park et al.}

\institute{$^1$University of Michigan \quad
$^2$Seoul National University \quad
$^3$NAVER AI Lab}

\maketitle

\begin{abstract}
With the extensive use of vision-language models in various downstream tasks, evaluating their robustness is crucial.
In this paper, we propose a benchmark for assessing the robustness of vision-language models. 
We believe that a robust model should properly understand both linguistic and visual semantics and be resilient to explicit variations.
In pursuit of this goal, we create new variants of texts and images in the MS-COCO test set and re-evaluate the state-of-the-art (SOTA) models with the new data. 
Specifically, we alter the meaning of text by replacing a word, and generate visually altered images that maintain some visual context while introducing noticeable pixel changes through image mixing techniques.
%
Our evaluations on the proposed benchmark reveal substantial performance degradation in many SOTA models (e.g., Image-to-Text Recall@1: 81.9\% $\rightarrow$ 48.4\% in BLIP, 66.1\% $\rightarrow$ 37.6\% in VSE$\infty$), with the models often favoring the altered texts/images over the original ones. 
This indicates the current vision-language models struggle with subtle changes and often fail to understand the overall context of texts and images.
Based on these findings, we propose semantic contrastive loss and visual contrastive loss to learn more robust embedding.
Datasets and code are available at {\url{https://github.com/pseulki/rococo}}.
\vspace{-1mm}
\vspace{-3mm}
\end{abstract}
\vspace{-3mm}
\section{Introduction}
\label{sec:intro}

\begin{figure*}[t]
\centering
\begin{subfigure}{0.66\linewidth}
    \includegraphics[width=1\linewidth]{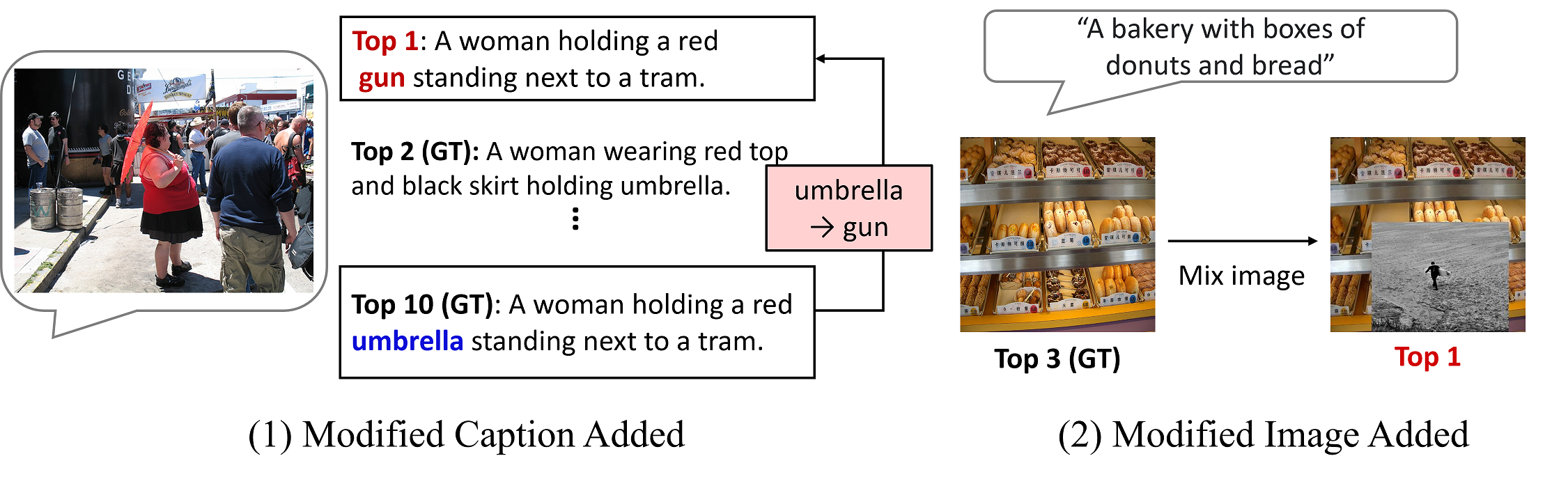}
    \caption{\textbf{{Example tested with BLIP~\cite{ref:li_blip_2022}}.}}
   \label{fig:blip_example}
\end{subfigure}
\hfill
\begin{subfigure}{0.33\linewidth}
    \includegraphics[width=1\linewidth]{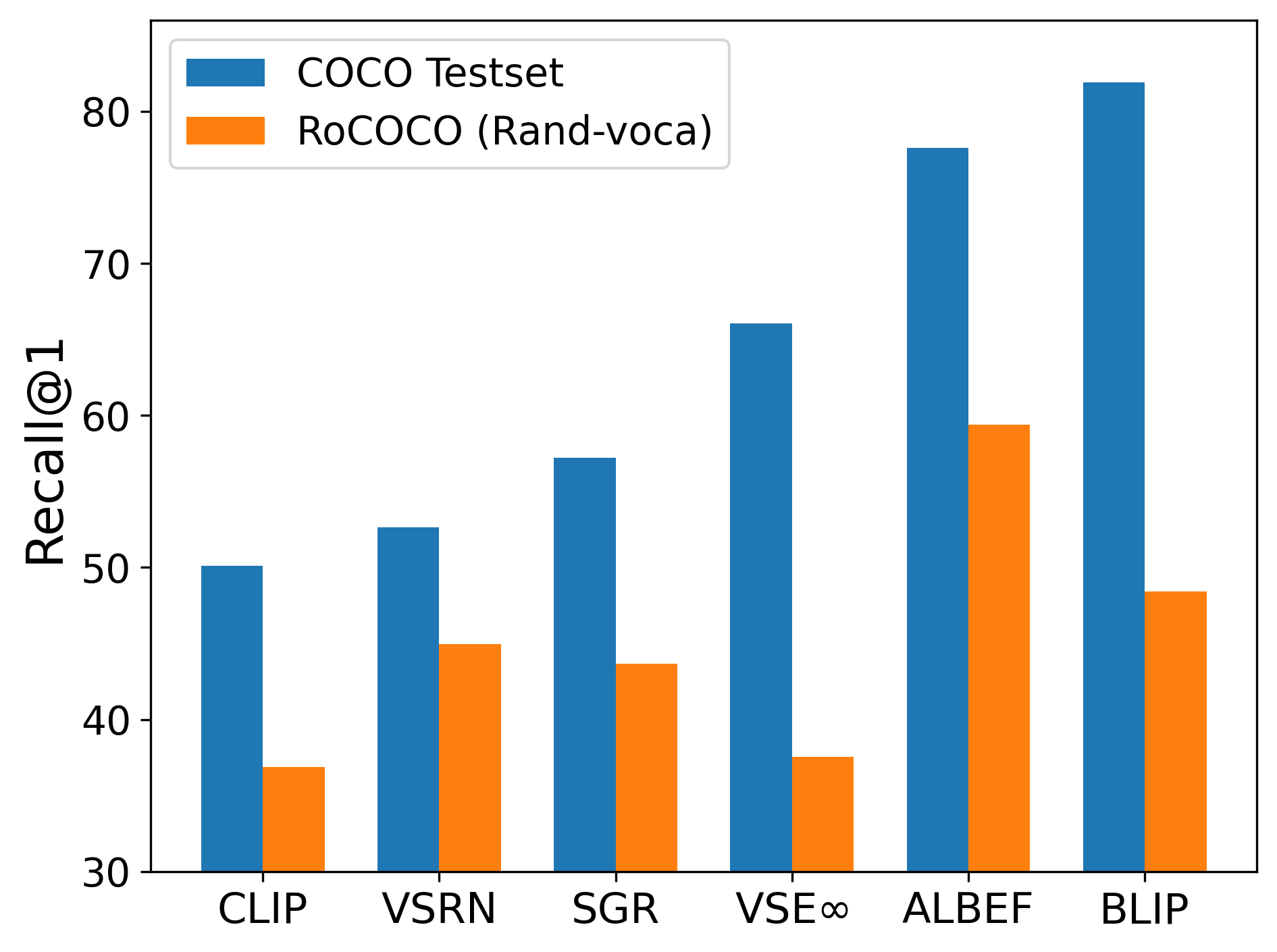}
    \caption{\textbf{Recall@1 Scores.}}
    \label{fig:intro_decrease}
\end{subfigure}
\vspace{-2mm}
\caption{\textbf{Motivating Example.} (a) When we add a new caption with only one word changed from ``umbrella'' to ``gun'', this new caption is preferred by the model (Image-to-text).
   Likewise, when we add a new image created by inserting an unrelated image to the original one, this new image is ranked as top 1 (Text-to-image). 
(b) Recall@1 scores of existing SOTAs decrease when tested with our RoCOCO benchmark.
}
  \label{fig:intro}
\vspace{-5mm}
\end{figure*}

Recently, vision-language (VL) models~\cite{ref:clip_2021, ref:align_2021} have generated significant research interest, with their learned visual-semantic features being actively employed across various domains.
For instance, CLIP~\cite{ref:clip_2021} latent features are used in  image generation models~\cite{ref:ramesh2022hierarchical, ref:galatolo2021generating, patashnik2021styleclip}, and CLIP scores are used to evaluate the alignment between generated images and texts~\cite{ref:saharia_neurips2022, ref:2022stylegan-nada}. 
Similarly, BLIP-Diffusion~\cite{ref:li2023blipdiffusion} uses a pre-trained module from BLIP-2~\cite{ref:li2023blip2} for controllable text-to-image generation.
Additionally, pretrained CLIP models are used for semantic understanding in robotic manipulation~\cite{ref:shridhar2022cliport}. 
Given the extensive use of VL models in various downstream tasks, evaluating the robustness of the models is imperative.

One of the primary methods for evaluating the VL models is through image-text matching (retrieval) tasks, which aim to align embedding features of images and texts effectively.
It is worth noting that the above-mentioned applications rely on image-text matching, which involves assessing the similarity between image features and text features.
Many vision-language models assess image-text matching accuracy (\emph{i.e.,} recall@1) on MS-COCO~\cite{ref:data_coco_lin_2014} or Flickr30K~\cite{ref:data2015flickr30k} benchmark datasets, with recent advancements showcasing notably strong performance~\cite{ref:li_vsrn_2019, ref:chun_pcme_2021, ref:chen_vse_infty_2021, ref:wang2022coder_eccv, ref:Kim_2023_CVPR, ref:zhang_vinvl_2021, ref:li_blip_2022}.
However, there are still important questions remaining: How much can we trust these numbers? 
How robust is it when used in the wild?

While existing evaluation datasets are meticulously curated, they often lack the diversity and variability inherent in real-world data. 
Thus, it is challenging to assess whether a model robustly understands semantic and visual details using existing datasets.
Figure~\ref{fig:blip_example} shows a motivating example of the retrieval results, when adding variants of texts and images to MS-COCO~\cite{ref:data_coco_lin_2014} test set, tested with the state-of-the-art (SOTA), BLIP~\cite{ref:li_blip_2022}.
When we retrieve the top-ranked text from a given image, the caption \textit{``A woman holding a red \textbf{\textcolor{red}{gun}} standing next to a tram''} is preferred the most over the original caption \textit{``A woman holding a red \textbf{\textcolor{blue}{umbrella}} standing next to a tram''} by the model. 
Likewise, given the text, the image partially mixed with a completely unrelated image (\emph{i.e.,} skiing on the snow) is retrieved as the highest similarity (top 1), instead of the correct clean image, which is ranked as top 3. 

Although these variants are easily distinguishable by humans, we are surprised to find that SOTA models can be misled by such  simple and explicit alterations.
Considering the widespread use of image-text feature similarity across many downstream tasks, the lack of robustness and the neglect of such details can lead to various issues.
If the highest text-image similarities are inaccurate, CLIP-based image generation evaluations~\cite{ref:saharia_neurips2022, ref:2022stylegan-nada} might lose credibility, and there is a risk of incorrect robotic execution~\cite{ref:shridhar2022cliport} due to a misunderstanding of commands.
Lastly, these misleading results can greatly damage the trust in AI systems.

From this motivation, we propose a Robustness benchmark of MS-COCO (RoCOCO) to stress-test image-text matching models. 
\textbf{Our initial belief is that a robust model should properly understand both linguistic and visual semantics and be resilient to explicit variations in texts or images. }
To explore this, we generate new sentences with minimal structural changes (\emph{i.e.,} altering only one word) while changing its meaning (\emph{e.g.,} ``umbrella'' $\rightarrow$ ``gun''), as our language can generate an infinite number of different sentences by changing just one word.
Similarly, for images, we contaminate the original images by mixing them with unrelated ones, thereby creating new images that retain some visual context but have disrupted visual coherence.
This synthetic approach allows us to create controlled variations that can stress-test the model's ability to handle diverse inputs.

\begin{figure*}[t]
\begin{center}
    \includegraphics[width=0.9\linewidth]{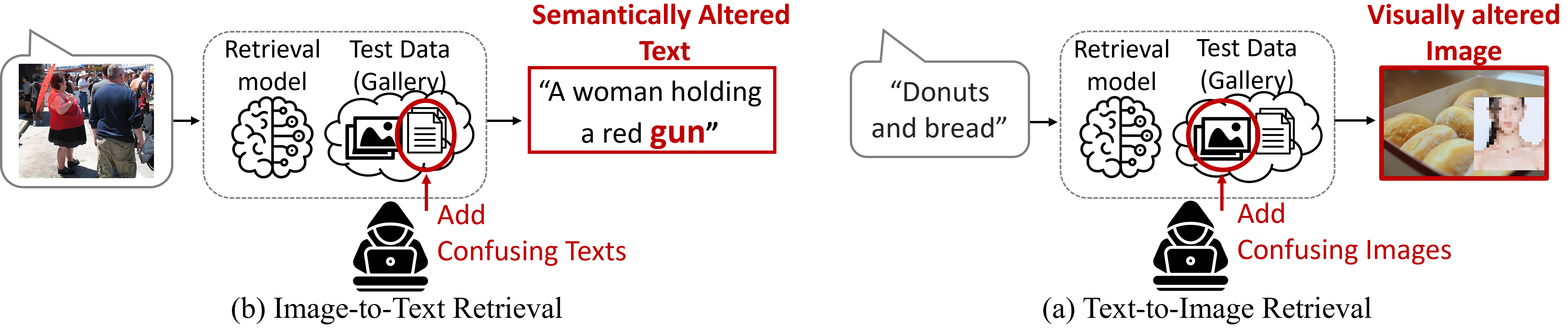}
  \vspace{-5mm}
\end{center}
   \caption{\textbf{Our Benchmark}. By adding newly created confusing texts and images into the existing test set, our proposed benchmarks stress-test the model's robustness in understanding visual and semantic details.
   }
\label{fig:intuition}
\vspace{-7mm}
\end{figure*}


These synthesized captions and images are added to the existing test set to evaluate the model's capability to accurately retrieve the original ground-truth captions and images without being confused by the additional contents (See Figure~\ref{fig:intuition}).
Despite the simplicity of the alteration, many state-of-the-art models show considerable performance degradation on the proposed benchmarks (\emph{e.g.,} $81.9\% \rightarrow 48.4\%$ in BLIP (ViT-B)~\cite{ref:li_blip_2022}, $66.1\% \rightarrow 37.6\%$ in VSE$\infty$ (WSL Grid)~\cite{ref:chen_vse_infty_2021} for Image-to-Text retrieval), as can be seen in Figure~\ref{fig:intro_decrease}.
Our discovery reveals \textbf{that models trained on even hundreds of millions of data points can be susceptible to errors which are easily discernible by human observers}.
In addition, our further analysis highlights the tendency of current image-text matching models to overlook subtle details and show more attention to specific words or image parts.

Finally, to address the vulnerabilities in understanding subtle details of image-text matching models, we propose Semantic Contrastive (SC) loss and Visual Contrastive (VC) loss.
The SC loss encourages the model to learn semantic details by increasing the embedding distance between an original image or caption and a randomly corrupted caption.
Similarly, the VC loss facilitates the effective learning of comprehensive visual information by enlarging the embedding distance between a randomly corrupted image and an original image or caption. 

Our key contributions can be summarized as follows:
\begin{itemize}[leftmargin=*,noitemsep,topsep=0pt]
\item[$\bullet$] We propose a robustness-evaluation benchmark for the image-text matching task, and discover significant performance drops across all models regardless of the extent of large-scale pre-training.
\item[$\bullet$] We study vulnerabilities of image-text matching models and observe that these models often tend to focus on specific words or image components rather than comprehending the overall context.
\item[$\bullet$] We propose Semantic and Visual Contrastive Losses to learn details in image and captions, enhancing the robust alignment in the embedding space.
\end{itemize}





\section{Related Work}
\vspace{-2mm}
\subsection{Image-Text Matching}
\vspace{-2mm}
\textbf{Methods.}
Most image-text matching (ITM) methods~\cite{ref:frome2013devise, ref:scan_2018, ref:song_pvse_2019, ref:cvse_2020, ref:noisy_2021_CVPR, ref:sgraf_2021,  ref:chun_pcme_2021, ref:wasserstein_retrieval_2021_iccv, ref:chen_vse_infty_2021} aim to learn joint visual-semantic embedding (VSE) such that paired image and text representation in the embedding space are close. 
%
In recent years, large-scale pre-training models~\cite{ref:uniter_2020, ref:oscar_2020, ref:zhang_vinvl_2021, ref:align_2021, ref:vilt_kim_2021, ref:albef_2021, ref:li_blip_2022, ref:cots_retrieval_2022_cvpr, ref:vista_cvpr_2022, ref:coca_2022, ref:flamingo_2022, ref:byun2022grit} have shown strong achievement in both zero-shot and fine-tuned performances.
Most of these models adopt transformer architecture and can learn cross-modal representations benefiting from large-scale image-text pairs. 

\noindent\textbf{Datasets.}
New ITM benchmark datasets, such as Crisscrossed Captions (CxC)~\cite{ref:crisscrossed_2020} and ECCV caption~\cite{ref:eccvcaption_2022}, have been proposed to extend MS-COCO. 
These datasets aim to improve associations and address false negatives in MS-COCO.
However, our main focus differs as we aim to assess the vulnerability of models rather than providing improved benchmark datasets.

\vspace{-2mm}
\subsection{Robustness Test}
\vspace{-2mm}

\textbf{Unimodal.}
Robustness in deep learning methods has been actively studied in both computer vision and natural language processing (NLP) areas.
In computer vision, one research direction is data poisoning~\cite{ref:biggio_poison_2012, ref:steinhardt2017certified, ref:hendrycks_noise_2018, ref:gu2017badnets, ref:chen2017targeted}, which attacks the robustness of models during training by adding images with small perturbations.
Meanwhile, adversarial attack studies ~\cite{ref:goodfellow2014explaining, ref:kurakin2016adversarial, ref:carlini2017adversarial, ref:sparseattack2019, ref:guo_attack_2019} inject imperceptible noises to test images so that a model can make wrong predictions.
For image retrieval task, Li et al.~\cite{ref:imageretrievalattack_2019} showed that adding invisible noise to a query image can make the model return incorrect images. 
Another line of research~\cite{ref:imagenet-c_2019, ref:imagenet-a_2021, ref:imagenet-r_2021} has proposed new ImageNet benchmarks for common robustness evaluation.
In NLP, research on data poisoning~\cite{ref:wallace2021concealed} and adversarial attacks~\cite{ref:2018hotflip, ref:alzantot2018generating, ref:jia2019certified, ref:garg2020bae, ref:li2021contextualized, ref:dong_nlp_wordsubst_2021, ref:boucher2022bad} has also been actively studied to fool the prediction of models. 
Adversarial examples are produced by character-level modifications~\cite{ref:belinkov2018synthetic}, paraphrasing sentences~\cite{ref:iyyer2018adversarial}, or substituting a word with a synonym~\cite{ref:ren2019generating, ref:li2020bert-attack}.
The main difference between these methods and our work is that while the previous works generate human imperceptible noises and texts that preserve their original meaning, we intentionally create images and texts that are perceptibly different and disrupt their original meaning.
Our intuition is straightforward: a robust model should not be confused by such simple alterations, which are too obvious for humans.


\noindent\textbf{Multimodal.}
As vision-language models have generated growing research interest, robustness work for cross-modal domain has been actively studied~\cite{ref:park_CLEVR_2019, ref:li2020closer, ref:carlini2022poisoning, ref:yuksekgonul2023when}.
Especially, in visual question answering (VQA), diverse robustness-evaluation benchmarks~\cite{ref:zhang_yinyang_2016, ref:goyal_vqa2_2017, ref:shah_vqarephrase_2019, ref:gokhale_vqalol_2020, ref:sheng_advqa_2021, ref:li_adversvqa_2021} have been proposed.
Recently proposed LANCE~\cite{ref:prabhu2023lance} shares similarity with ours in terms of manipulating words within text and introducing variations in images. 
LANCE generates counterfactual images with diverse transformations using a language guide by employing image-text models, such as CLIP~\cite{ref:clip_2021} and BLIP~\cite{ref:li_blip_2022}.
Our purpose is different from LANCE as we aim to create a benchmark dataset for evaluating those models, CLIP and BLIP.

\section{Robustness-Evaluation Benchmark}
\vspace{-1mm}
\subsection{Observations motivating the proposed approach}
Our goal is to quantitatively evaluate how well vision-language (VL) models understand both text and image.
Specifically, we 
measure the robustness of a VL model through our proposed benchmark, which assesses how robustly the model retrieves the ground-truth image or caption 
instead of 
our visually altered image or semantically altered caption.

Based on the examples observed from the BLIP~\cite{ref:li_blip_2022} model from Figure~\ref{fig:blip_example}, we generate altered captions and images that are 
capable of assessing the model's vulnerability.
Firstly, to assess the model's ability for understanding the overall details between the image and text, we introduce semantically altered  captions to make the image-to-text task challenging. 
Specifically, we create a semantically altered caption by replacing one word in the caption to change the meaning of the sentence. 
For example, replacing ``umbrella'' with ``gun'' as in Figure~\ref{fig:blip_example} (1). 

%

Likewise, we generate a visually altered image with noticeable changes by mixing an unrelated image into an original image (Figure~\ref{fig:blip_example} (2)). 
Surprisingly, even though the visually altered image is easily discernible by humans, we observe that the VL model often favors a mixture of unintended images rather than the desired (ground-truth) ones.
%
In the following sections, we describe the generation process in detail.


\vspace{-1mm}
\subsection{Semantically Altered  Caption Generation}
\subsubsection{Source Word Selection via Embedding-Influence.}\label{subsub:ei_source}
To introduce discernable changes in the meaning of a caption,
we focus on nouns for replacement.
For effective confusion in a caption, we 
choose a word that has minimal impact on the embedding output.
This idea is inspired by the common practice in which models measure similarity between the feature embedding of image and text encoders trained on image-text pairs~\cite{ref:clip_2021, ref:chen_vse_infty_2021, ref:li_blip_2022}. 
We hypothesize that even with considerable changes in the semantic meaning, the model would be confused with the original (ground-truth) caption if the feature embedding changes little.

To estimate the influence of a word, we propose embedding-influence (EI) score.
EI sore measures the change in embedding when the word is removed from the caption.
Given a text encoder $f_T$, and a caption  $C=\{c_m ~|~m=1, \cdots, M\}$, where $M$ is the number of words in $C$, the embedding-influence (EI) score of a word, $c_s$, is defined by
\vspace{-2mm}
\begin{equation}\label{eq:ei_score}
    EI(c_s) = 1 - \frac{<f_T(C),  f_T(C\setminus c_s)>}{\parallel f_T(C)\parallel \parallel f_T(C\setminus c_s)\parallel},
\end{equation}
where $<,>$ denotes the dot (inner) product operation.
A low EI score means that the word has little influence on the embedding output of the caption.
Given its limited influence compared to other words, substituting this word with a different word is expected to have low impact on the overall embeddings, which may lead to confusision between the modified caption and the original (ground-truth) caption.
We demonstrate this choice in Section~\ref{appendix:sec:inf_word}.


\begin{table*}[t]
\caption{\textbf{Added Concept Groups.}
We include new concepts. Table shows added unique concepts and 3 random words from each group. Examples from Same-concept can test if the model can correctly recognize the semantic differences.
}
\vspace{-2mm}
\label{table:concept_group}
\centering
\small
\resizebox{0.9\linewidth}{!}
{
\begin{tabular}{lcll}
\toprule
concept group & \#concepts & concept lemmas (sampled)  & examples from `Same-concept' (\textbf{Ground-truth} / \textbf{\textcolor{red}{Semantically-altered}})  \\ \midrule
material      & 32         & metal, plastic, wooden &  Five bagels are on a (\textbf{metal} / \textbf{\textcolor{red}{wood}}) rack.    \\
color         & 28         & black, white, brown   &   (\textbf{Brown} / \textbf{\textcolor{red}{Green}}) and black dog looking at a person holding a frisbee.    \\
direction     & 50         & front, top, bottom   & A book sitting on (\textbf{top} / \textbf{\textcolor{red}{bottom}}) of a wooden desk.     \\
vehicle\_part & 12         & hood, wheel, tire     &   A cat by an overturned pot and a bicycle (\textbf{wheel} / \textbf{\textcolor{red}{timer}}).    \\
shape         & 15         & round, square, octagon  & (\textbf{Square}/ \textbf{\textcolor{red}{Round}}) dishes hold main dishes while a banana is ...    \\
event         & 11         & Christmas, birthday, wedding & A table set for a traditional (\textbf{Thanksgiving} / \textbf{\textcolor{red}{Christmas}}) dinner. \\
number        & 14         & one, five, hundreds    &  (\textbf{Hundreds} / \textbf{\textcolor{red}{Five}}) of people gathered in the park with ...    \\ \bottomrule
\end{tabular}
}
\vspace{-0.4cm}
\end{table*}




\begin{wraptable}{r}{5.5cm}
\vspace*{-4mm}
\caption{\textbf{Models for EI scores.}
}
\label{table:ei-models}
\centering
\scriptsize
{
\begin{tabular}{lcl}
\toprule
Model  & COCO-trained & Text backbone \\ \midrule
VSRN & Finetuned  & Bi-GRU~\cite{ref:cho_gru_2014}           \\
CLIP & Zero-shot  & Transformer~\cite{ref:vaswani_transformer_2017}   \\
VSE$\infty$  & Finetuned  & BERT~\cite{ref:bert_2019}          \\
BLIP & Finetuned  & BERT~\cite{ref:bert_2019}          \\ \bottomrule
\end{tabular}
\vspace*{-0.8cm}
}
\end{wraptable}

To ensure generalizability and not limit the choice of nouns to a specific model, we use four representative models in Table~\ref{table:ei-models} (i.e., VSRN~\cite{ref:li_vsrn_2019}, CLIP~\cite{ref:clip_2021}, VSE$\infty$~\cite{ref:chen_vse_infty_2021}, BLIP~\cite{ref:li_blip_2022}), and measure the EI score of each word. 
We select a word with the least influence across the models. If the word is chosen by the majority of models, it is replaced by a target word (see Section \ref{target word}).
If there are multiple candidates of nouns, we randomly choose one. 
Interestingly, the words with the lowest embedding influence exhibit little variation across the models. 
We provide further details in Appendix Section~\ref{appendix:sec:modelvariation}.
In our evaluation, we not only include these four models but also add ten additional models not involved in noun selection process for a more comprehensive assessment.


 


\vspace{-3mm}
\subsubsection{Altered Word Selection for Diverse Benchmark Captions.}
\label{target word}
To generate semantically altered captions covering various scenarios, we need to determine an ``altered (replacement) word'' replacing the source (original) word chosen in the previous section. To choose an altered word, we employ four different policies.
First, we use concept groups from GRIT benchmark~\cite{ref:gupta_grit_2022}, which categorizes nouns from popular datasets including COCO into 24 concept groups such as food, people, and places.
We add 7 concept groups (see Table~\ref{table:concept_group}) for words that are hard to be included in the given concept groups.
Next, we categorize words according to the concept groups.
Then, an original  word is randomly replaced by any altered word inside \textbf{Same-concept} or  \textbf{Diff-concept}.
For example, \textbf{Same-concept} replaces ``umbrella'' with an altered word from the same concept (i.e., tools), which can be ``rope'' or ``boxes''. 
\textbf{Diff-concept} replaces ``umbrella'' with an altered word selected randomly from different concepts, such as ``pizza'' from ``food'' concept, or ``monkey'' from ``animal'' concept. 

\begin{figure*}[t]
\centering
   \includegraphics[width=0.9\linewidth]{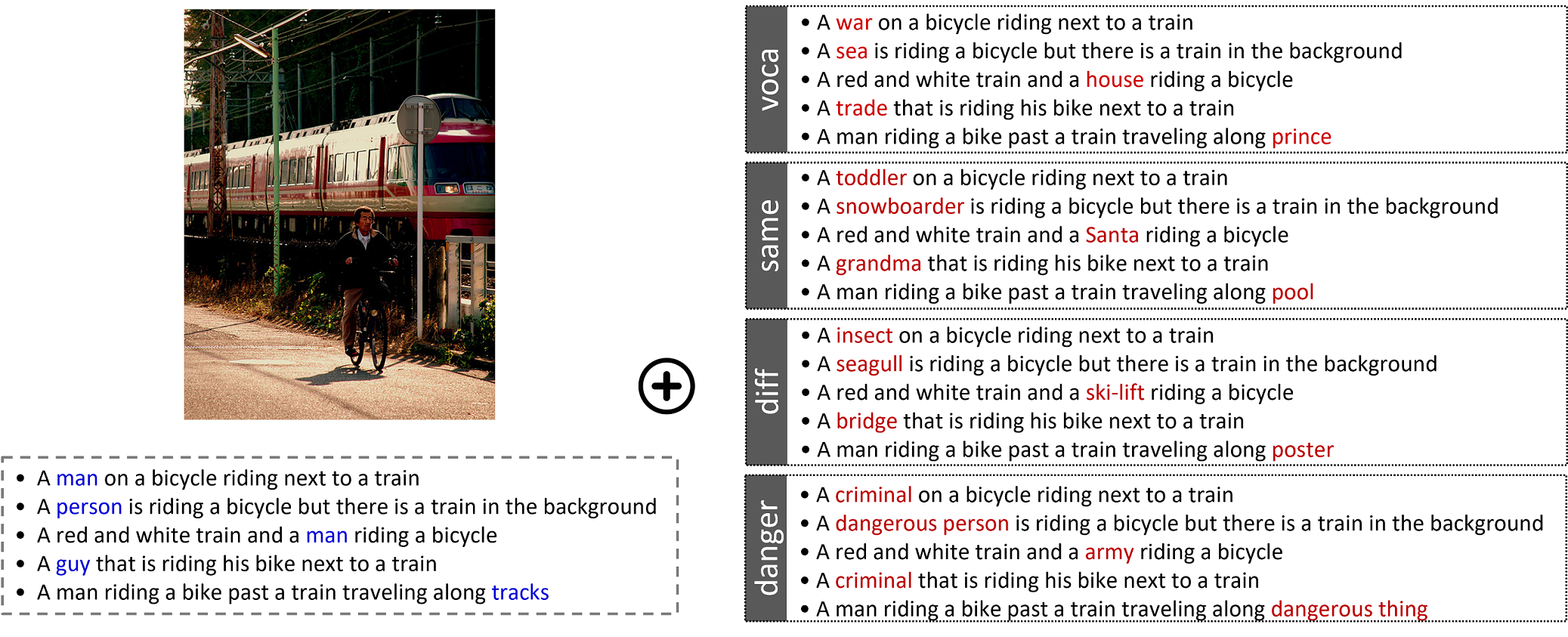}
   \vspace{-2mm}
   \caption{\textbf{Example of semantically altered captions.}
   (Left) Original MS-COCO image and captions. (Right) Our generated captions, Rand-voca, Same-concept, Diff-concept, and Danger from top to bottom. 
   The model's robustness is evaluated if it can correctly retrieve the original \blue{ground-truth caption}, in the presence of newly generated \red{semantically altered captions}.
   }
\label{fig:voca_ex}
\vspace{-5mm}
\end{figure*}

Next, to stress-test with a wider range of words, we employ the BERT~\cite{ref:bert_2019} vocabulary (\textbf{Rand-voca}). 
We randomly select an altered word consisting of only English letters, excluding those in other languages or special characters. 
Lastly, we create a special case (\textbf{Danger}) by using words related to public security. 
This allows us to evaluate the models' ability to comprehend critical situations that could potentially pose a threat to human safety. 
For instance, we replace ``umbrella'' with ``gun'' or ``weapon''. 
Examples of the semantically altered captions can be seen in Figure~\ref{fig:voca_ex}.


\begin{figure}[b]
\begin{center}
\includegraphics[width=0.6\columnwidth]{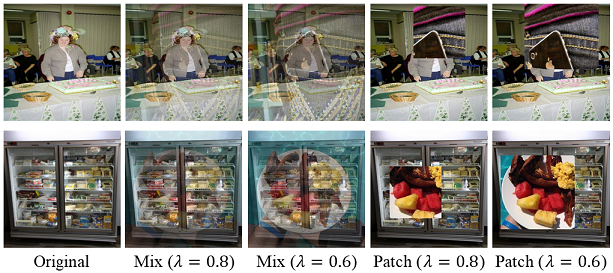}
\vspace{-2mm}
\caption{\textbf{Example of visually altered images with different $\lambda$.}}
\label{fig:mixed_examples}
\end{center}
\vspace{-8mm}
\end{figure}

\subsection{Visually Altered Image Generation}
\vspace{-1mm}
To generate visually altered images that preserve some visual context but have disrupted visual coherence, we employ two simple image mixing techniques. 
One is the Mixup-style approach~\cite{ref:mixup_iclr2018}, where two images are blended together in different proportion (Mix). 
The other method inserts a patch of an unrelated image 
onto the original image, as in CutMix~\cite{ref:cutmix_iccv2019} (Patch).
For simplicity, the unrelated image is randomly selected from the COCO test set, but it can be selected from any other dataset.
When inserting an unrelated image $x^f$ into an original image $x^o$, we use two 
mixing ratios $\lambda$ and $\textbf{M}$ for Mix and Patch, respectively, as follows:
\begin{align*}
\vspace{-2mm}
  \text{Mix} &:  \Tilde{x} = \lambda x^o + (1 - \lambda) x^f, \\
  \text{Patch} &: \Tilde{x} = \textbf{M} \odot x^o + (\textbf{1} - \textbf{M}) \odot x^f, \nonumber
    \label{eq:mix}
\vspace{-2mm}
\end{align*}
where $\textbf{M} \in \{0, 1\}^{W \times H}$ denotes a binary mask at a randomly chosen location of the unrelated patch, where $W$ is  width and $H$ is  height of an image.
The portion of 1 in M is adjusted according to the mixing ratio of  $\lambda = \frac{\sum_{i,j} \textbf{M}_{i,j}}{W \times H}$.
Figure~\ref{fig:mixed_examples} shows the examples of visually altered images. 
Creating these visually altered images and adding them to the gallery set provides an easy yet effective method to measure the robustness of the model.


\section{Experiments and Results}
\subsection{Experimental setting}
In this section, we evaluate the existing image-text matching (ITM) models on our new dataset, RoCOCO. 
For Image-to-Text retrieval, we expand MS-COCO test data~\cite{ref:karpathy2015coco} by adding new 25,000 semantically altered captions to the existing 25,000 original captions, creating a gallery of 50,000 captions. 
We then retrieve text from this expanded gallery. Likewise, for Text-to-Image retrieval, we add new 5,000  visually altered images to the existing 5,000 original images, resulting in an image gallery of 10,000 images.

\textbf{Evaluation Metrics.}
Recall@k, especially Recall@1 (R@1), is the most popular metric for evaluating the existing ITM methods.
In this paper, we propose two metrics, \textit{Drop Rate} and {\textit{Recall Score of Manipulated Samples} (RSMS)} in addition to R@1.
Drop rate measures the relative decrease in R@1 compared to the evaluation on the original COCO 5K test set.
We calculate drop rate as $(\text{R}@1-\text{R}_{\text{New}}@1) / \text{R@1}$.
RSMS calculates the percentage of newly added semantically altered  captions and visually altered images that are retrieved as top 1.
This can quantitatively estimate the vulnerability of a model.

\textbf{Models for Evaluation.}
We compare 14 state-of-the-art Image-Text Matching (ITM) models, whose trained weights are available to the public.
They can be categorized into two groups; large-scale vision-language (VL) pre-training and visual semantic embedding  groups. 
Large-scale VL pre-training group 
includes CLIP with ViT-B/32, ViT-B/16 and ViT-L/14 backbones~\cite{ref:clip_2021}, fine-tuned ALBEF~\cite{ref:albef_2021}, and zero-shot and fine-tuned BLIP with ViT-B and ViT-L backbones~\cite{ref:li_blip_2022}.
While `zero-shot' and `fine-tuned' models are both pre-trained on large-scale datasets, `zero-shot' refers to not being fine-tuned with COCO train set. 
Visual semantic embedding group includes models using region features based on bottom-up attention~\cite{ref:bottomup_2018} and SCAN~\cite{ref:scan_2018}: VSRN~\cite{ref:li_vsrn_2019}, SAF, SGR~\cite{ref:sgraf_2021}, and VSE$\infty$~\cite{ref:chen_vse_infty_2021}.

\subsection{Re-evaluation on RoCOCO}
\subsubsection{Image-to-Text Retrieval.}
Table~\ref{table:main-i2t} reports the image-to-text retrieval results on our new datasets.
First, we observe the highest performance degradation on Rand-voca, with BLIP ViT-B showing a decrease of nearly 50\%.
This can be attributed to the fact that Rand-voca contains numerous unexpected words that do not commonly appear together in captions. 
This observation suggests that models are 
vulnerable to sentences comprising unfamiliar word combinations that rarely appear in the trained captions. 

\begin{table*}[t]
\caption{\textbf{Image-to-Text retrieval results.}
Models are re-evaluated on our newly proposed RoCOCO datasets: Rand-voca, Same-concept, Diff-concept, and Danger.
Recall@1 (R@1)($\uparrow$), drop rate($\downarrow$), Recall Score of Manipulated Samples (RSMS)($\downarrow$) are shown. We can observe consistent degradation across all vision-language models. 
}
\label{table:main-i2t}
\centering
\small
\resizebox{1\linewidth}{!}
{
\begin{tabular}{lclccclccclccclccc}
\toprule
                          & COCO 5K &  & \multicolumn{3}{c}{Rand-voca} &  & \multicolumn{3}{c}{Same-concept} &  & \multicolumn{3}{c}{Diff-concept} &  & \multicolumn{3}{c}{Danger} \\ \cline{2-2} \cline{4-6} \cline{8-10} \cline{12-14} \cline{16-18} 
                          & R@1           &  & R@1     & drop rate   & RSMS  &  & R@1      & drop rate    & RSMS   &  & R@1      & drop rate    & RSMS   &  & R@1    & drop rate  & RSMS \\ \midrule
\multicolumn{18}{l}{{Large-scale VL pre-training models}}                                                                                                                            \\
CLIP ViT-B/32 (zero-shot)~\cite{ref:clip_2021} & 50.10         &  & 36.88   & \red{-26.39\%}       & 33.56      &  & {36.58}    & \red{-26.99\%}        &   {34.52}     &  & 38.28    & \red{-23.59\%}        &   31.18     &  & 42.18  & \red{-15.81\%}      &   19.69   \\
CLIP ViT-B/16 (zero-shot)~\cite{ref:clip_2021} & 52.44         &  & {40.16}   & \red{-{23.42}\%}       &   {31.26}    &  &    39.84      &  \red{-24.03\%}            &  31.68      &  &   40.56       &  \red{-22.65\%}            &  28.52      &  &   44.67     &    \red{-14.81\%}        &   18.19   \\
CLIP ViT-L/14 (zero-shot)~\cite{ref:clip_2021} & 56.04         &  & {41.42}   & \red{-{26.09}\%}       &  31.72     &  &  43.08        &   \red{-23.13\%}           &  {30.06}      &  &    43.38      &   \red{-22.59\%}           &     28.24   &  &   46.48     &     \red{-17.06\%}       &   30.16   \\
ALBEF~\cite{ref:albef_2021}                   & 77.58         &  & {59.38}   & \red{-{23.46}\%}       &  {26.76}     &  &    60.92      &   \red{-21.47\%}           &   24.94     &  &  60.44        &   \red{-22.09\%}           &  24.22      &  &   63.37     &    \red{-18.32\%}        &   20.43   \\
BLIP ViT-B (zero-shot)~\cite{ref:li_blip_2022}    & 70.54         &  & {34.84}   & \red{-{50.61}\%}       &  {55.62}     &  & 48.52         &   \red{-31.22\%}           &    37.26    &  &      43.30    &     \red{-38.62\%}         &    43.92    &  &    42.39    &      \red{-39.90\%}      &  43.99    \\
BLIP ViT-B~\cite{ref:li_blip_2022}                & 81.90         &  & {48.44}   & \red{-{40.85}\%}       &  {44.06}     &  & 59.30    & \red{-27.59\%}        &     31.42   &  & 57.28    & \red{-30.06\%}        &  34.14      &  & 67.81  & \red{-17.21\%}       &  18.92    \\
BLIP ViT-L (zero-shot)~\cite{ref:li_blip_2022}    & 73.66         &  & {47.34}   & \red{-{35.73}\%}       &    {38.62}   &  &      56.98    &       \red{-22.64\%}       &    26.68    &  &     55.68     &         \red{-24.41\%}     &    27.78    &  &    55.93    &        \red{-24.07\%}    &    26.54  \\
BLIP ViT-L~\cite{ref:li_blip_2022}                & 82.36         &  & {64.56}   & \red{-{21.61}\%}      &    {24.00}   &  &     70.46     &     \red{-14.45\%}         &   16.88     &  &    70.32      &       \red{-14.62\%}       &     17.00   &  &   72.37     &   \red{-12.13\%}         &  13.73    \\ \midrule
\multicolumn{18}{l}{{Visual Semantic Embedding models}}                                                                                                               \\
VSRN~\cite{ref:li_vsrn_2019}                      & 52.66         &  & {44.98}   & \red{-{14.58}\%}       & {16.28}      &  & 47.38     & \red{-10.03\%}        &   12.32     &  & 48.36    & \red{-8.17\%}         &   9.42     &  & 46.78  & \red{-11.17\%}      &   12.77   \\
SAF~\cite{ref:sgraf_2021}                       & 55.46         &  & {41.32}   & \red{-{25.50}\%}       &    {28.46}   &  & 44.70    & \red{-19.40\%}        &    24.02    &  & 47.50    & \red{-14.35\%}        &   18.28     &  & 42.77  & \red{-22.88\%}       &   26.35   \\
SGR~\cite{ref:sgraf_2021}                       & 57.22         &  & {43.66}   & \red{-{23.70}\%}       &   {26.86}    &  & 45.68    & \red{-20.17\%}        &    23.78    &  & 48.00    & \red{-16.11\%}         &     19.46   &  & 44.90  & \red{-21.53\%}      &  24.72    \\
VSE$\infty$ (BUTD region)~\cite{ref:chen_vse_infty_2021}         & 58.02         &  & {32.72}   & \red{-{43.61}\%}       &   {46.76}    &  & 41.78    & \red{-27.99\%}        &    31.60    &  & 37.90    & \red{-34.68\%}        &   38.72     &  & 37.66  & \red{-35.09\%} &   37.38   \\
VSE$\infty$ (BUTD grid)~\cite{ref:chen_vse_infty_2021}           & 59.40         &  & {31.78}   & \red{-{46.50}\%}       &   {48.92}    &  & 42.36    & \red{-28.69\%}        &   31.68     &  & 39.00    & \red{-34.34\%}        &   38.40     &  & 39.71  & \red{-33.15\%}      &  35.32    \\
VSE$\infty$ (WSL grid)~\cite{ref:chen_vse_infty_2021}            & 66.06         &  & {37.56}   & \red{-{43.14}\%}       &   {46.08}    &  & 50.00    & \red{-24.31\%}        &   27.74     && 45.08    & \red{-31.76\%}        &   34.32     &  & 45.39  & \red{-31.29\%}       &   33.07   \\ \bottomrule
\end{tabular}
\vspace*{-15mm}
}
\end{table*}

\begin{figure*}[t]
\begin{center}
\includegraphics[width=1\linewidth]{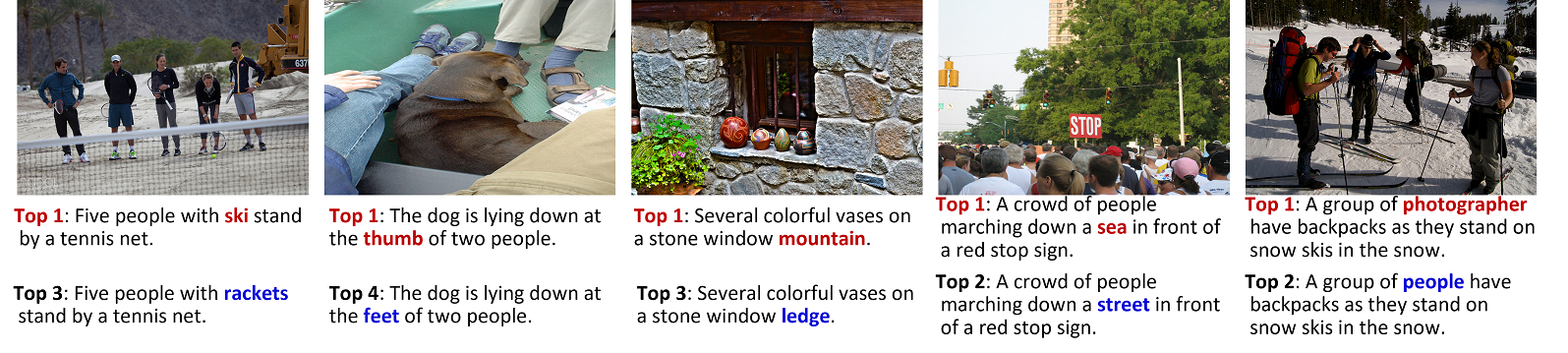}
\vspace{-5mm}
\caption{\textbf{Examples of incorrectly retrieved texts with BLIP from Same-concept (Image-to-Text).} 
\red{Captions with incorrect details} are ranked at the top 1, while \blue{the correct captions} are ranked lower.
}
\label{fig:i2t_example}
\end{center}
\vspace{-8mm}
\end{figure*}

\begin{table*}[t]
\caption{\textbf{Text-to-Image retrieval.}
Models are evaluated with our new Mix and Patch with different $\lambda$. 
We report Recall@1 (R@1)($\uparrow$), drop rate($\downarrow$), Recall Score of Manipulated Samples (RSMS)($\downarrow$), averaged over three random image generations with different seeds.
We can see consistent degradation across all vision-language models. 
}
\label{table:main-t2i}
\centering
\small
\resizebox{1\linewidth}{!}
{
\begin{tabular}{lclccclccclccclccc}
\toprule
                          & COCO 5K &  & \multicolumn{3}{c}{Mix ($\lambda=0.9$)} &  & \multicolumn{3}{c}{Mix ($\lambda=0.8$)} &  & \multicolumn{3}{c}{Patch ($\lambda=0.9$)} &  & \multicolumn{3}{c}{Patch ($\lambda=0.8$)} \\ \cline{2-2} \cline{4-6} \cline{8-10} \cline{12-14} \cline{16-18} 
                          & R@1     &  & R@1        & drop rate      & RSMS      &  & R@1        & drop rate      & RSMS     &  & R@1         & drop rate       & RSMS      &  & R@1         & drop rate       & RSMS      \\ \midrule
\multicolumn{18}{l}{{Large-scale VL pre-training models}}                                                                                                                                                               \\
CLIP ViT-B/32 (zero-shot)~\cite{ref:clip_2021} & 30.14   &  & 20.29      & \red{-32.68\%}          &   33.55        &  & 22.79      & \red{-24.39\%}          &   26.03        &  & 22.49       & \red{-25.38\%}           &  28.63         &  & 24.15       & \red{-19.87\%}           &   23.69        \\
CLIP ViT-B/16 (zero-shot)~\cite{ref:clip_2021} & 33.03   &  & 20.05      & \red{-39.30\%}          &   39.00        &  & 23.57      & \red{-{28.64}\%}          &   29.88        &  & 22.58       & \red{-{31.64}\%}           &    35.18       &  & 24.70       & \red{-{25.22}\%}           & {29.41}          \\
CLIP ViT-L/14 (zero-shot)~\cite{ref:clip_2021} & 36.14   &  & 25.49      & \red{-29.47\%}          &   28.99        &  & 27.75      & \red{-23.22\%}          &    24.29       &  & 27.56       & \red{-23.74\%}           &    27.64       &  & 29.09       & \red{-19.51\%}           &    23.97       \\
ALBEF~\cite{ref:albef_2021}                     &  60.67       &  &       44.13     &          \red{-27.27\%}      &     26.60      &  &  48.02          &      \red{-20.85\%}          &   21.11        &  &    48.86         &     \red{-19.47\%}            &   19.58        &  &   51.80          &            \red{-14.62\%}     &      15.30     \\
BLIP ViT-B (zero-shot)~\cite{ref:li_blip_2022}    & 56.36   &  & 39.03      & \red{-30.75\%}          &     31.54      &  & 43.94      & \red{-22.04\%}          &    22.28       &  & 41.96       & \red{-25.55\%}            &    27.56       &  & 45.05       & \red{-20.07\%}            &    22.79       \\
BLIP ViT-B~\cite{ref:li_blip_2022}                & 64.31   &  & 40.71      & \red{-{36.70}\%}           &   39.93        &  & 46.97      & \red{-26.96\%}          &    30.84       &  & 48.40       & \red{-24.74\%}           &   {42.57}        &  & 52.61       & \red{-18.19\%}           &     21.45      \\
BLIP ViT-L (zero-shot)~\cite{ref:li_blip_2022}    &  58.18       &  &     44.29       &      \red{-23.87\%}          &         25.13  &  &    47.61        &       \red{-18.17\%}         &    19.96       &  &   46.79          &      \red{-19.58\%}           &    21.07       &  &    49.50         &      \red{-14.93\%}           &    16.50       \\
BLIP ViT-L~\cite{ref:li_blip_2022}               &  65.06       &  &    41.87        &        \red{-35.64\%}        &         {42.45}  &  &    48.92        &         \red{-24.81\%}       &    {33.91}       &  &      48.55       &       \red{-25.38\%}          &    29.17       &  &         49.50    &      \red{-23.92\%}           &       22.10    \\ \midrule
\multicolumn{18}{l}{{Visual Semantic Embedding models}}                                                                                                                                               \\
VSRN~\cite{ref:li_vsrn_2019}                      & 40.34   &  & 27.04      & \red{-32.97\%}          &   39.05        &  & 31.36      & \red{-{22.26}\%}          &  {28.87}         &  & 30.08       & \red{-{25.43}\%}           &       {31.11}    &  & 32.50       & \red{-{19.43}\%}           &     {24.80}      \\
SAF~\cite{ref:sgraf_2021}                       & 40.11   &  & 30.90      & \red{-22.96\%}          &   27.84        &  & 33.37      & \red{-16.80\%}          &  22.87         &  & 32.50       & \red{-18.97\%}           &   23.78        &  & 34.03       & \red{-15.16\%}           &     19.69      \\
SGR~\cite{ref:sgraf_2021}                      & 40.45   &  & 30.71      & \red{-24.08\%}           &    28.08       &  & 33.41      & \red{-17.40\%}           &    22.57       &  & 32.40       & \red{-19.90\%}            &      23.95     &  & 34.08       & \red{-15.75\%}           &     19.90      \\
VSE$\infty$ (BUTD region)~\cite{ref:chen_vse_infty_2021}         & 42.46   &  & 31.57      & \red{-25.65\%}          &  30.74         &  & 35.61      & \red{-16.13\%}          &     20.45      &  & 34.17       & \red{-19.52\%}           &  23.51         &  & 36.48       & \red{-14.08\%}           &      17.28     \\
VSE$\infty$ (BUTD grid)~\cite{ref:chen_vse_infty_2021}           & 44.07   &  & 30.22      & \red{-31.43\%}          &    36.68       &  & 35.26      & \red{-19.99\%}          &  25.00         &  & 35.70       & \red{-18.99\%}           &   23.52        &  & 38.75       & \red{-12.07\%}           &  15.82         \\
VSE$\infty$ (WSL grid)~\cite{ref:chen_vse_infty_2021}            & 51.55   &  & 34.31      & \red{-{33.44}\%}          &  {38.60}         &  & 40.40      & \red{-21.63\%}          &    26.26       &  & 43.67       & \red{-15.29\%}           &  18.39         &  & 46.87       & \red{-9.08\%}            &  11.31         \\ \bottomrule
\end{tabular}
\vspace*{-7mm}
}
\end{table*}

\begin{figure*}[t]
\begin{center}
\includegraphics[width=1\linewidth]{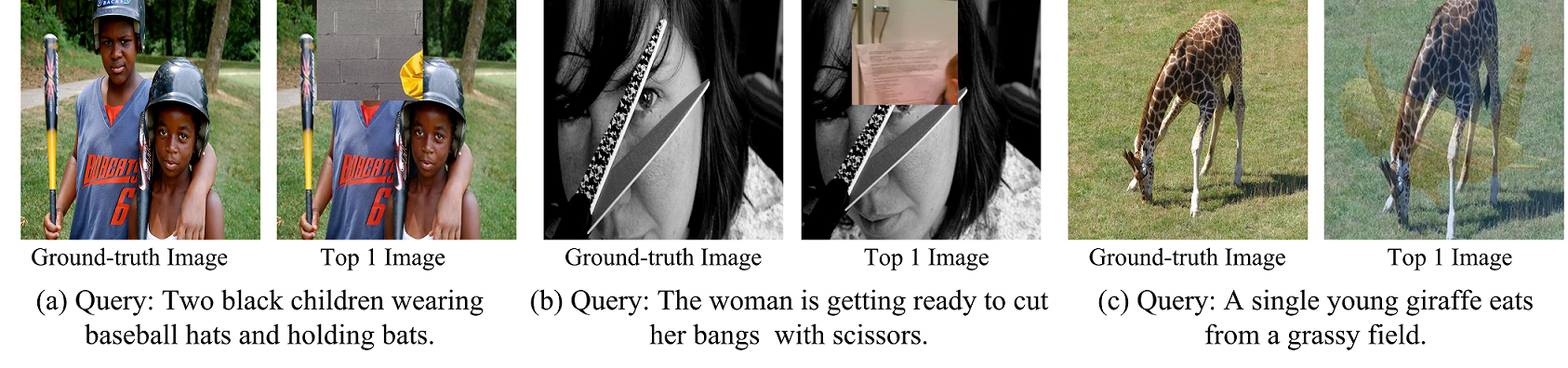}
\vspace*{-7mm}
\caption{\textbf{Examples of images incorrectly retrieved by BLIP  when $\lambda=0.8$ (Text-to-Image).} The first two examples are from the Patch, while the last one is from the Mix. 
In the Patch examples, some important areas are obscured. For instance, the face of the boy on the left is obscured, which is a crucial detail for relating to the query ``Two black children.''
In the Mix example, unrelated image of a ``plane'' is visible.
}
\label{fig:t2i_example}
\end{center}
\vspace{-7mm}
\end{figure*}

Furthermore, we can observe consistent degradation across all vision-language models, with drop rates ranging from 11\% to 51\% and RSMS from 13\% to 55\%, regardless of 
methods or the scale of pre-training datasets (e.g., 400M image pairs in CLIP~\cite{ref:clip_2021}, 129M in BLIP~\cite{ref:li_blip_2022}, 14M in ALBEF~\cite{ref:albef_2021}).
We assume that commonly used image-text matching loss might be vulnerable to a single-word change in the caption because the loss is used to minimize only the distance between image-text pairs for learning multimodal representations. 
In addition, Figure~\ref{fig:i2t_example} presents qualitative examples evaluated with BLIP (ViT-B) from Same-concept dataset.
Our results highlight the importance of developing a robust training strategy for ITM models that can better capture word-level semantic meaning and align it with images.

\vspace{-3mm}
\subsubsection{Text-to-Image Retrieval.}
We evaluate VL methods on the new image set with $\lambda=0.9, 0.8$ in Table~\ref{table:main-t2i}.
The images are generated using three random seeds, and the averaged results are reported. 
It can also be observed that all VL methods consistently exhibit performance degradation, with drop rates ranging from 9\% to 39\% and RSMS from 11\% to 42\%.

In addition, in Figure~\ref{fig:t2i_example}, we present examples of incorrect image retrievals using BLIP (ViT-B) when $\lambda$ is set to 0.8. 
For instance, the face of the boy on the left is obscured, which is a crucial detail for relating to the query ``Two black children.''
While humans would generally prefer the ground-truth clean images over these mixed images, we observe that the models often overlook details and prefer the contaminated images. 
%
We argue that this evaluation is simple yet effective for assessing the robustness of the models. 
More results with different $\lambda$ values can be found in Appendix (Section~\ref{appendix:sec:t2i}).

\vspace*{-1mm}
\subsection{Analysis and Discussions}\label{subsec:Findings}

\subsubsection{Each word within a caption has a different impact on the embedding.}\label{appendix:sec:inf_word}
\begin{wrapfigure}{r}{6cm}
\vspace*{-5mm}
\centering
{
\includegraphics[width=1\linewidth]{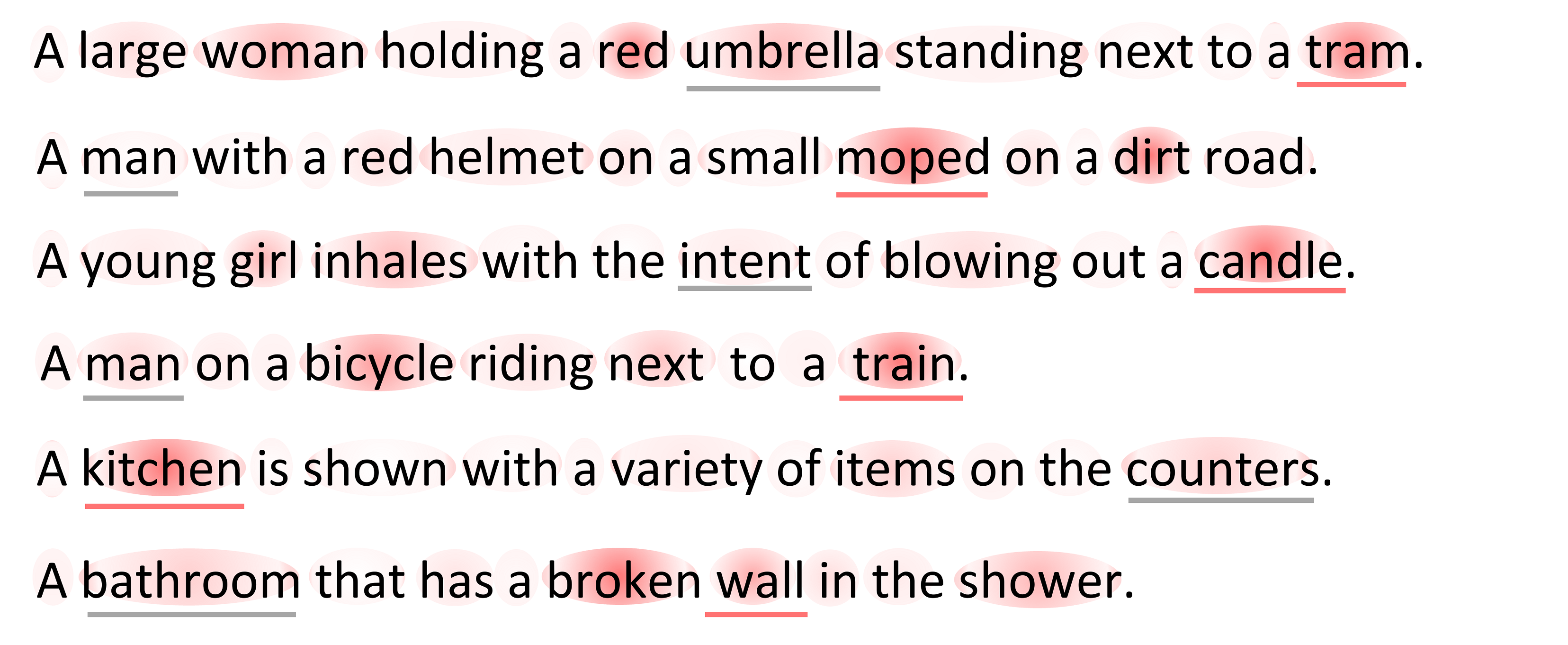}
\vspace*{-5mm}
  \caption{\textbf{Varying influence of a word in a caption.} Red indicates higher influence.}\label{fig:inf_example}
}
\vspace*{-5mm}
\end{wrapfigure}

In Section~\ref{subsub:ei_source}, we introduce the Embedding-Influence (EI) score. 
Figure~\ref{fig:inf_example} illustrates the varying EI scores of words within each caption, with the red color 
indicating higher influence. 
The noun with the highest EI score is underlined in red, while the noun with the lowest score is underlined in gray. 
For instance, in the first caption, the noun ``umbrella'' has a relatively lower influence compared to the noun ``tram,'' even though it carries significant meaning within the caption. 
Therefore, replacing the word ``umbrella'' can lead to a notable change in semantic meaning without significantly affecting the original embedding.

\vspace{-2mm}
\subsubsection{Manipulating words with low EI scores proves to be an effective approach for semantic-shift confusion.}\label{subsec:low_ei}
To demonstrate this, we evaluate model performance by removing words in captions using different methods. 
``Random'' method randomly removes a noun, while  ``Large EI'' removes the noun with the highest EI score, and vice versa for ``Low EI''.
We create new captions by simply deleting the original word without replacement to mitigate the impact of the altered word. 
Table~\ref{table:ablation-ei} shows that deleting words with low EI scores is the most effective approach for misleading the models, while deleting words with high EI scores results in minimal performance degradation. 
This finding supports our hypothesis that leveraging the influence of words on embedding features can effectively confuse the models. 
Thus, manipulating words with low EI scores can be an effective method for assessing the robustness of newly trained models.

 \begin{table*}[t]
\caption{\textbf{Effects of using EI scores.}
Deleting a source word with the lowest EI score results in the largest performance drop, causing the feature embedding to remain confusingly similar to the original.
}
\vspace{-3mm}
\label{table:ablation-ei}
\centering
\small
\resizebox{1\linewidth}{!}
{
\begin{tabular}{lccccccccccccc}
\toprule
                          & COCO  &  & \multicolumn{3}{c}{Random Deletion} &  & \multicolumn{3}{c}{High EI Deletion} &  & \multicolumn{3}{c}{Low EI Deletion} \\ \cline{2-2} \cline{4-6} \cline{8-10} \cline{12-14} 
                          & R@1($\uparrow$)   &  & R@1($\uparrow$)       & drop rate($\downarrow$)    & RSMS($\downarrow$)     &  & R@1($\uparrow$)       & drop rate($\downarrow$)     & RSMS($\downarrow$)      &  & R@1($\uparrow$)       & drop rate($\downarrow$)     & RSMS($\downarrow$)      \\ \midrule
CLIP ViT-B/32 (zero-shot)~\cite{ref:clip_2021} & 50.10 &  & 38.58     & \red{-22.99\%}        & 29.66    &  & 42.76     & \red{-14.65\%}         & 21.84     &  & \textbf{36.04}     & \red{-\textbf{28.06}\%}         & \textbf{32.30}     \\
CLIP ViT-L/14 (zero-shot)~\cite{ref:clip_2021} & 56.04 &  & 42.54     & \red{-24.09\%}        & 30.4     &  & 48.58     & \red{-13.31\%}         & 20.42     &  & \textbf{39.22}     & \red{-\textbf{30.01}\%}         & \textbf{33.74}     \\
BLIP ViT-B (zero-shot)~\cite{ref:li_blip_2022}    & 70.54 &  &   45.58        &   \red{-35.38\%}      &  40.54    &  & 57.14          &   \red{-19.00\%}     &   25.80    &  &    \textbf{36.34}  &    \red{-\textbf{48.48}\%}      &   \textbf{52.48}    \\
BLIP ViT-B~\cite{ref:li_blip_2022}                & 81.90 &  & 65.54     & \red{-22.46\%}        & 19.98    &  & 72.74     & \red{-11.18\%}         & 14.06     &  & \textbf{59.28}     & \red{-\textbf{27.62}\%}         & \textbf{30.10}     \\
VSRN~\cite{ref:li_vsrn_2019}                      & 52.66 &  & 44.7      & \red{-15.12\%}        & 18.02    &  & 43.46     & \red{-17.47\%}         & 22.56     &  & \textbf{38.56}     & \red{-\textbf{26.78}\%}         & \textbf{29.36}     \\
VSE$\infty$ (BUTD region)~\cite{ref:chen_vse_infty_2021}         & 58.02 &  & 34.2      & \red{-41.05\%}        & 45.58    &  & 40.58     & \red{-30.06\%}         & 38.06     &  & \textbf{30.02}     & \red{-\textbf{48.26}\%}         & \textbf{50.72}     \\
VSE$\infty$ (BUTD grid)~\cite{ref:chen_vse_infty_2021}           & 59.40 &  & 34.3      & \red{-42.26\%}        & 46.46    &  & 39.92     & \red{-32.79\%}         & 39.78     &  & \textbf{30.46}     & \red{-\textbf{48.72}\%}         & \textbf{51.54}     \\
VSE$\infty$ (WSL grid)~\cite{ref:chen_vse_infty_2021}            & 66.06 &  & 40.8      & \red{-38.24\%}        & 41.68    &  & 47.32     & \red{-28.37\%}         & 33.76     &  & \textbf{36.56}     & \red{-\textbf{44.66}\%}         & \textbf{47.14}     \\ \bottomrule
\end{tabular}
\vspace{-4mm}
}
\end{table*}
\begin{figure*}[t!]
\begin{center}
\includegraphics[width=1\linewidth]{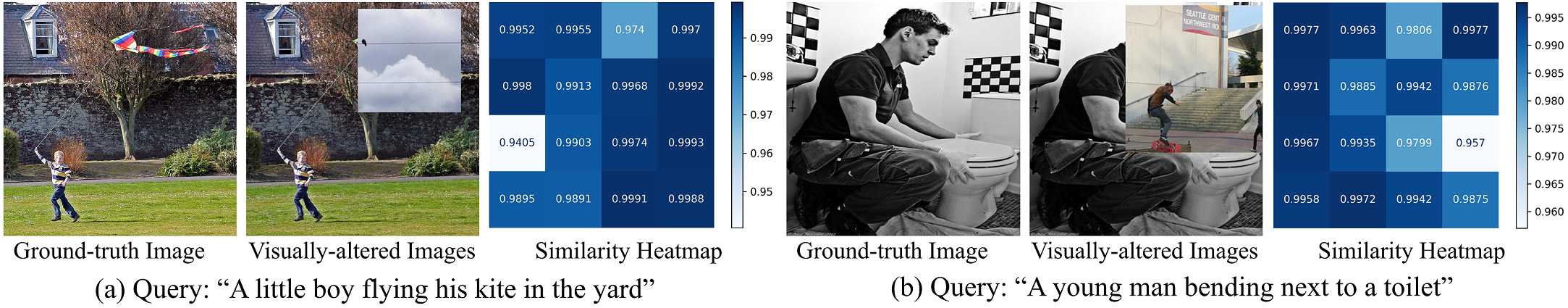}
\vspace*{-4mm}
\caption{\textbf{Varying influence of spatial regions on the embedding.}
The similarity heatmap shows the cosine similarity between the embedding of the original image and the embedding when each part is masked. Dark colors indicate higher similarity (less influence), meaning the embedding changes little when that part is removed. White colors indicate important areas where significant changes occur.
Even when specific parts are replaced with unrelated scenes (e.g., a kite), the model may overlook these changes because other, more influential parts (e.g., the boy's face) remain.
}
\label{fig:image_embedding_analysis}
\end{center}
\vspace*{-5mm}
\end{figure*}

\vspace{-2mm}
\subsubsection{The influence of each spatial part on the embedding varies within a single image.}
To examine why the model can be deceived by unrelated images, we analyze the impact of each spatial location on the feature embedding. 
We divide the image into 16 parts and mask each part to zeros, to observe the changes in the embedding. 
The heatmap in Figure~\ref{fig:image_embedding_analysis} illustrates the cosine similarity between the embedding of the original image and 
the image embedding when each corresponding part is masked.
In the cases where visually altered images are retrieved as top 1, we can observe that influential parts like ``boy'' or ``toilet'' still remain despite obscuring some important parts like ``kyte'' or ``a man's face''.
This finding indicates that certain parts of the image have a stronger impact on the retrieval outcomes than the other parts.


\begin{table*}[t!]
\caption{\textbf{Image-to-Text retrieval on dataset with multiple words substitutions.}
The results are averaged over generations with three different random seeds.
Recall@1 (R@1)($\uparrow$), drop rate($\downarrow$), Recall Score of Manipulated Samples (RSMS)($\downarrow$) are shown. 
Models can confuse sentences even when semantic meaning is largely damaged.
}
\vspace{-3mm}
\label{table:abl-multi}
\centering
\small
\resizebox{1\linewidth}{!}
{
\begin{tabular}{lclccclccclccclccc}
\toprule
                          & COCO  &  & \multicolumn{3}{c}{2 words substitution} &  & \multicolumn{3}{c}{3 words substitution} &  & \multicolumn{3}{c}{4 words substitution} &  & \multicolumn{3}{c}{5 words substitution} \\ \cline{2-2} \cline{4-6} \cline{8-10} \cline{12-14} \cline{16-18} 
                          & R@1   &  & R@1       & drop rate       & RSMS       &  & R@1       & drop rate       & RSMS       &  & R@1       & drop rate       & RSMS       &  & R@1       & drop rate       & RSMS       \\ \midrule
\multicolumn{18}{l}{\textbf{Large-scale VL pre-training models}}                                                                                                                                                             \\
CLIP ViT-B/32 (zero-shot)~\cite{ref:clip_2021} & 50.10 &  &    42.89       &     \red{-14.39\%}            &      19.71      &  &     46.07      &        \red{-8.04\%}         &      12.67      &  &     47.45      &     \red{-5.29\%}            &     8.15       &  &   48.37        &        \red{-3.45\%}         &       5.46     \\
CLIP ViT-B/16 (zero-shot)~\cite{ref:clip_2021} & 52.44 &  &     45.35      &        \red{-13.52\%}         &        19.07    &  &     48.43      &        \red{-7.65\%}         &      11.89      &  &     49.97      &              \red{-4.71\%}   &       8.01     &  &     50.61      &        \red{-3.49\%}         &       5.95     \\
CLIP ViT-L/14 (zero-shot)~\cite{ref:clip_2021} & 56.04 &            &       47.35          &          \red{-15.51\%}  & 22.18 &           &        50.22         &       \red{-10.39\%}     & 15.78 &           &    51.99             &     \red{-7.23\%}       & 11.56 &           &    53.07   &     \red{-5.30\%}     &      8.27      \\
ALBEF~\cite{ref:albef_2021}                     & 77.58 &  &     72.43      &       \red{-6.64\%}          &        2.40    &  &     73.03      &      \red{-5.86\%}           &      0.88      &  &     73.23      &             \red{-5.61\%}    &      0.43      &  &     73.26      &        \red{-5.57\%}         &       0.32     \\
BLIP ViT-B (zero-shot)~\cite{ref:li_blip_2022}    & 70.54 &  &     53.04      &      \red{-24.81\%}           &     30.75       &  &    62.99       &        \red{-10.70\%}         &      14.72      &  &     67.95      &        \red{-3.67\%}         &      5.44      &  &     69.73      &         \red{-1.15\%}        &      1.86      \\
BLIP ViT-B ~\cite{ref:li_blip_2022}               & 81.90 &  &    73.62       &     \red{-10.11\%}            &       12.76     &  &   77.45        &         \red{-5.43\%}        &      7.10      &  &     79.54      &        \red{-2.88\%}         &       4.05     &  &     80.48      &    \red{-1.73\%}             &       2.51     \\
BLIP ViT-L (zero-shot)~\cite{ref:li_blip_2022}    & 73.66 &  &    60.35       &          \red{-18.07\%}       &      21.66      &  &      67.99     &           \red{-7.70\%}      &     10.16       &  &      71.63     &      \red{-2.76\%}           &    3.93        &  &      72.87     &          \red{-1.07\%}       &     1.61       \\
BLIP ViT-L ~\cite{ref:li_blip_2022}               & 82.36 &  &    73.93       &      \red{-10.24\%}           &   12.65         &  &    77.93       &   \red{-5.38\%}              &    7.45        &  &   79.81        &      \red{-3.10\%}           &       4.23     &  &     80.98      &     \red{-1.68\%}            &      2.54      \\ \hline
\multicolumn{18}{l}{\textbf{Visual Semantic Embedding models}}                                                                                                                                                \\
VSRN~\cite{ref:li_vsrn_2019}                      & 52.66 &  &    45.07       &        \red{-14.41\%}         &     17.79       &  &     47.89      &         \red{-9.06\%}        &   11.33         &  &    49.89       &      \red{-5.26\%}           &      7.08      &  &      50.99     &         \red{-3.17\%}        &    4.29        \\
SAF~\cite{ref:sgraf_2021}                       & 55.46 &  &     44.06      &      \red{-20.56\%}           &     20.29       &  &    47.22       &     \red{-14.86\%}            &    26.71        &  &     50.02      &   \red{-9.81\%}              &     15.12       &  &    51.71       &           \red{-6.76\%}      &     10.85       \\
SGR~\cite{ref:sgraf_2021}                       & 57.22 &  &     43.57      &      \red{-23.86\%}           &     28.53       &  &    46.98       &       \red{-17.90\%}          &     22.79       &  &      49.81     &     \red{-12.95\%}            &     17.49       &  &     51.91      &        \red{-9.28\%}         &     13.09       \\
VSE$\infty$ (BUTD region)~\cite{ref:chen_vse_infty_2021}         & 58.02 &  &    33.94       &     \red{-41.50\%}            &    46.81        &  &     37.15      &       \red{-35.98\%}          &       42.66     &  &    40.39       &       \red{-30.39\%}          &   37.79         &  &      43.17     &          \red{-25.60\%}       &      33.01      \\
VSE$\infty$ (BUTD grid)~\cite{ref:chen_vse_infty_2021}           & 59.40 &  &     34.79      &       \red{-41.44\%}         &  45.95         &  &    38.03       &     \red{-35.98\%}            &  41.75        &  &   41.17       &    \red{-30.68\%}             &      37.14     &  &    44.97       &      \red{-24.30\%}           &     30.57       \\
VSE$\infty$ (WSL grid)~\cite{ref:chen_vse_infty_2021}            & 66.06 &  &   39.95        &             \red{-39.52\%}    &       43.79     &  &    44.04       &               \red{-33.33\%}  &    38.44        &  &      48.29     &             \red{-26.90\%}    &      32.85      &  &      51.73     &           \red{-21.69\%}      &       27.51     \\ \bottomrule
\end{tabular}
}
\vspace{-0.2cm}
\end{table*}

\begin{figure*}[t]
\centering
\includegraphics[width=0.95\linewidth]{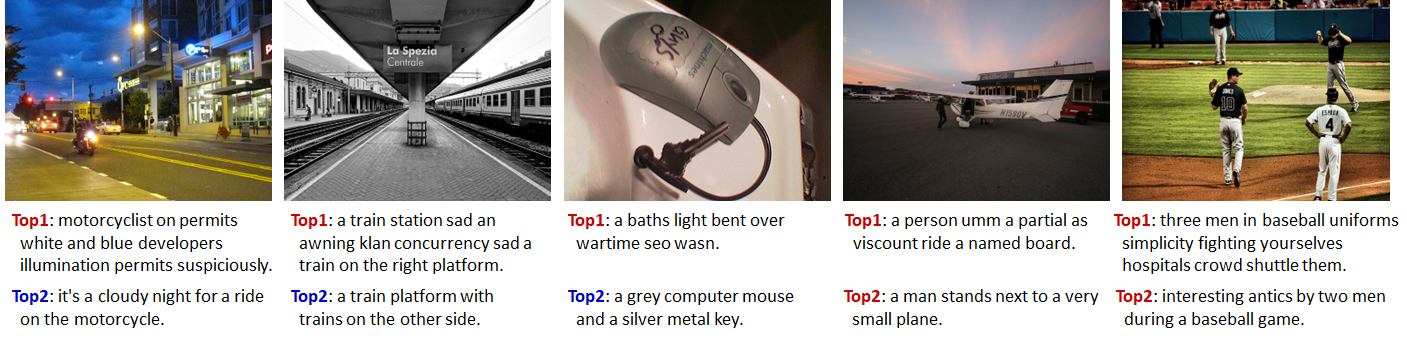}
\vspace{-0.1cm}
   \caption{\textbf{Example of substituting four random words evaluated with BLIP (ViT-B).}
We find that the model can be confused by highly nonsensical and fragmented sentences.
We hypothesize that the model focuses on specific words (e.g., ``motorcycle" in the first image) rather than comprehending the language as a whole.
   }
\label{fig:multi_ex}
\vspace*{-3mm}
\end{figure*}

\vspace{-2mm}
\subsubsection{VL models can be misled by highly nonsensical sentences with multiple word replacements.}
To further investigate the vulnerability of VL models, we conduct experiments where two to five words in the captions are randomly replaced with words from the BERT vocabulary.
Due to the brevity of some captions, we do not restrict the source words to nouns.
Figure~\ref{fig:multi_ex} presents examples of top-1 retrieval results from BLIP (ViT-B) for captions with four word replacements.
Interestingly, these broken captions often include at least one correct keyword, such as ``motorcyclist'' in the first image. 
This suggests that the model may prioritize specific words over understanding the sentence as a whole.
These findings are consistent with the observation by ~\cite{ref:yuksekgonul2023when}, that VL models behave like bags-of-words, ignoring word order in sentences. Our results with random word replacements further reveal that VL models may  heavily rely on the presence of specific words, even when the sentences are nonsensical and fragmented.
Additionally, the results in Table~\ref{table:abl-multi} quantitatively show meaningful performance degradation across the model, when the original semantic meaning is significantly disrupted.
Large-scale pretraining methods demonstrate better robustness than VSE models when multiple words are changed simultaneously.


\vspace{-2mm}
\section{Semantic and Visual Contrastive Loss for Robustness}\label{sub:defense}
\vspace{-2mm}
Throughout our study, we have observed that VL models tend to overlook semantic and visual details. 
To address this issue, we propose the Semantic Contrastive (SC) loss and {Visual Contrastive (VC) loss}, which encourages the model to learn details in texts and images, respectively.  
Given a text encoder $f_T$, an image encoder $f_I$, an image $x$, a corrupted image $x_{corrupt}$, a caption $c$, and a corrupted caption $c_{corrupt}$, SC and VC losses are defined by:
\begin{equation}
\vspace{-2mm}
\small
    L_{SC} = \frac{1}{2}\left(\underbrace{\frac{<f_T(c_{corrupt}),  f_I(x)>}{\parallel f_T(c_{corrupt})\parallel \parallel f_I(x)\parallel}}_{L(c_{corrupt}, x)} + \underbrace{\frac{<f_T(c_{corrupt}),  f_T(c)>}{\parallel f_T(c_{corrupt})\parallel \parallel f_T(c)\parallel}}_{L(c_{corrupt}, c)}\right),
\end{equation}
\begin{equation}
\small
    L_{VC} = \frac{1}{2}\left(\underbrace{\frac{<f_I(x_{corrupt}),  f_I(x)>}{\parallel f_I(x_{corrupt})\parallel \parallel f_I(x)\parallel}}_{L(x_{corrupt}, x)} + \underbrace{\frac{<f_I(x_{corrupt}),  f_T(c)>}{\parallel f_I(x_{corrupt})\parallel \parallel f_T(c)\parallel}}_{L(x_{corrupt}, c)}\right).
\vspace{1mm}
\end{equation}

Minimizing the losses will encourage the model to decrease the similarity (i.e., increase the difference) between the corrupted and original embeddings.
Firstly, to improve robustness for semantic details, SC loss has two components. (1) $L(c_{corrupt}, x)$ reduces the similarity between the embeddings of the original images and their corresponding corrupted captions.
This loss allows the model to capture shifts in meaning within similar sentences, with the aim of achieving a more solid alignment between the embeddings of the ground-truth images and text.
(2) $L(c_{corrupt}, c)$ encourages the model to increase the difference between similar sentences, thereby capturing semantic details more effectively.
Likewise, to learn visual details within images, VC loss has two components.
(1) $L(x_{corrupt}, x)$ is designed to increase the distance between corrupted image and original image embeddings. 
(2) $L(x_{corrupt}, c)$ increases the distance between corrupted image and original caption embeddings to enhance the robust alignment between caption and visual details.

In each batch, we generate a corrupted caption $c_{corrupt}$ by randomly selecting words within the caption to be replaced with a probability of $p$ (set to 0.3). 
These selected words are then substituted with random words from the BERT vocabulary with a probability of $q$ (set to 0.5), or deleted with a probability of $1-q$.
Corrupted images are created with a 0.5 probability by randomly manipulating images within a batch using either the Mixup~\cite{ref:mixup_iclr2018} or CutMix~\cite{ref:cutmix_iccv2019}.

\begin{figure*}[t!]
\centering
    \begin{subfigure}{0.24\linewidth}
    \includegraphics[width=\linewidth]{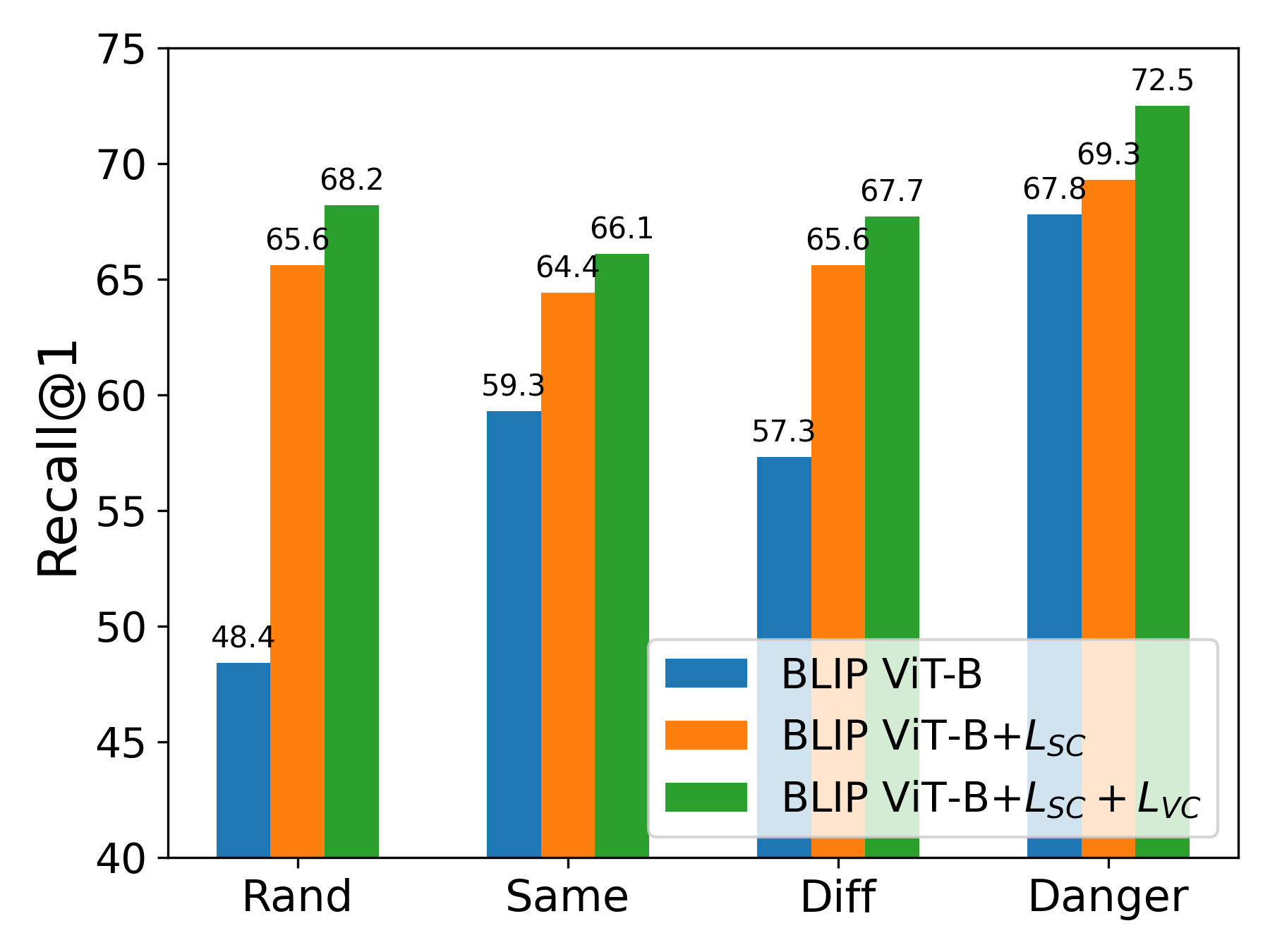}
    \caption{Recall@1($\uparrow$)}
    \end{subfigure}
    \begin{subfigure}{0.24\linewidth}
    \includegraphics[width=\linewidth]{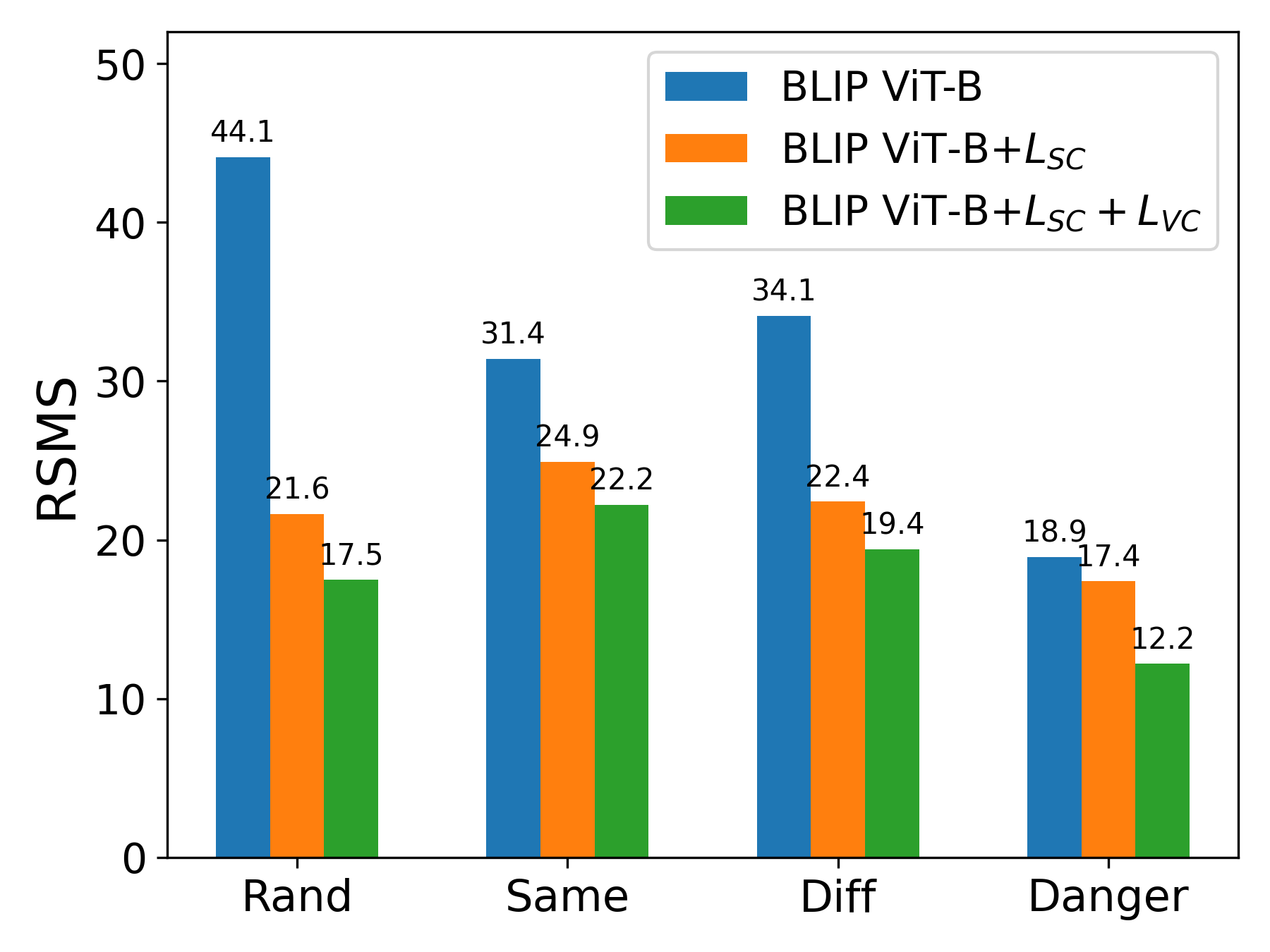}
    \caption{RSMS($\downarrow$)}
    \end{subfigure}
    \begin{subfigure}{0.24\linewidth}
    \includegraphics[width=\linewidth]{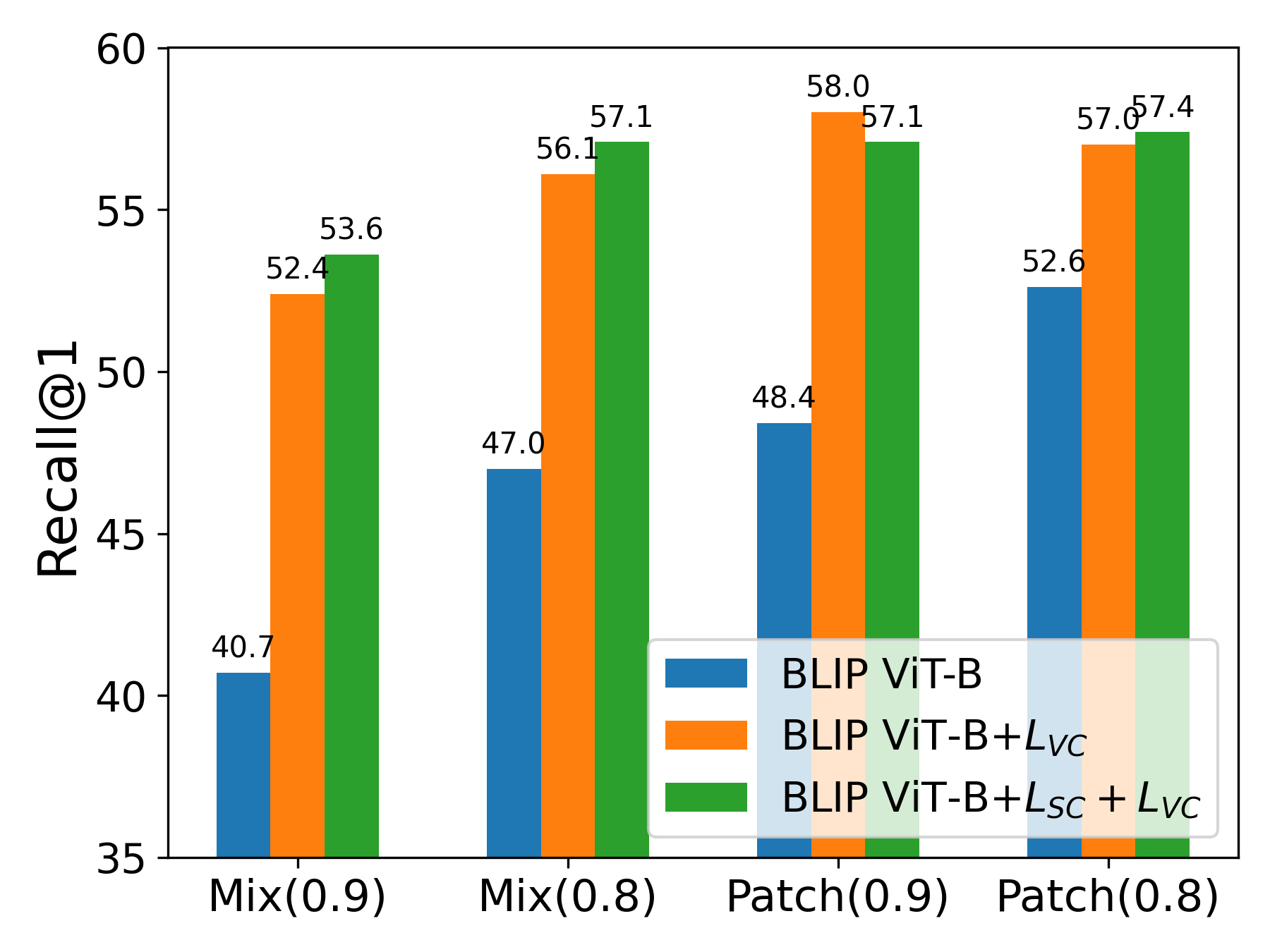}
    \caption{Recall@1($\uparrow$)}
    \end{subfigure}
    \begin{subfigure}{0.24\linewidth}
    \includegraphics[width=\linewidth]{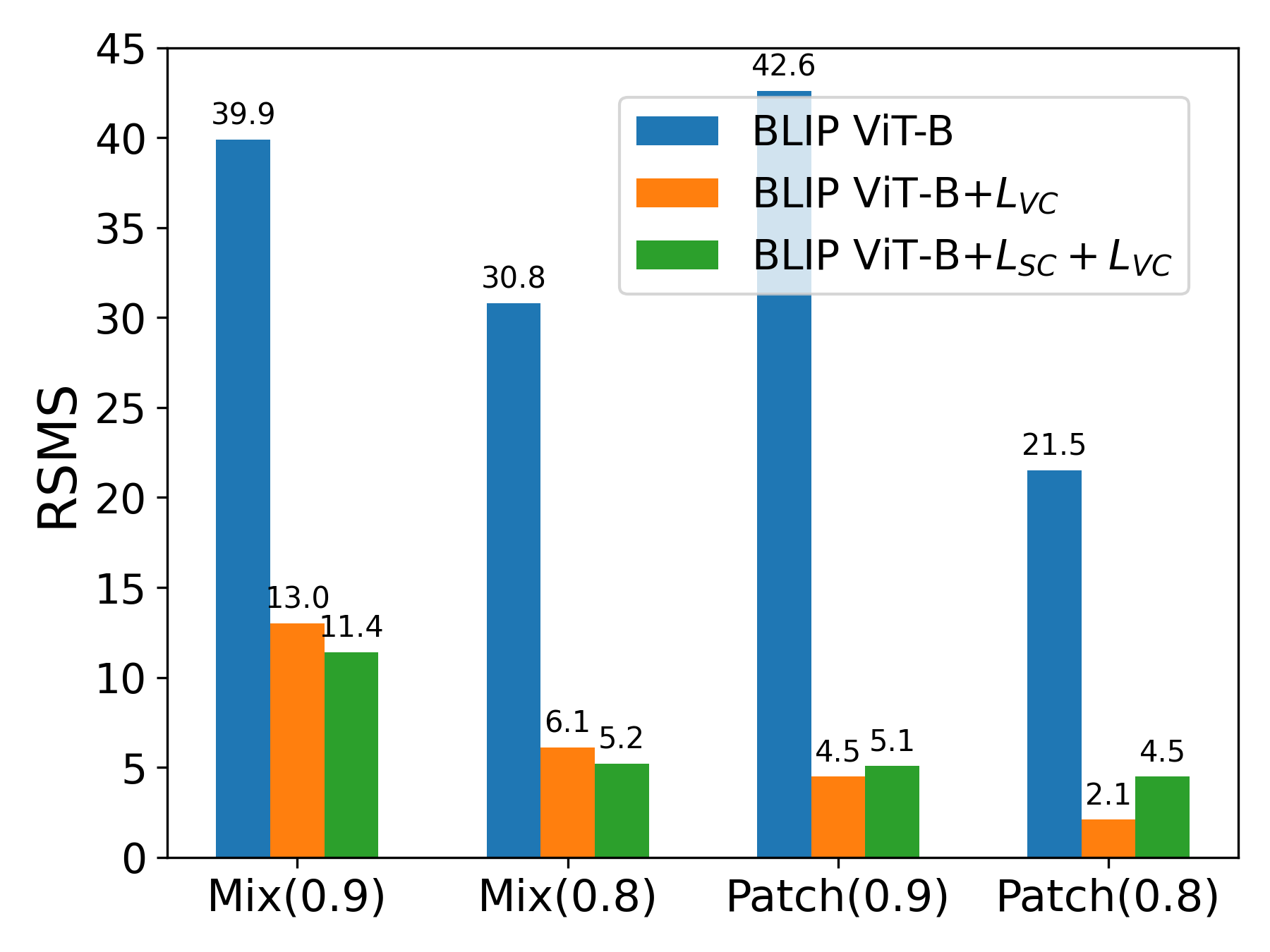}
    \caption{RSMS($\downarrow$)}
    \end{subfigure}
\caption{\textbf{Simple Semantic Contrastive Loss and Visual Contrastive Loss can significantly improve robustness of BLIP (ViT-B) model.} (a) and (b): Image-to-Text Retrieval. (c) and (d): Text-to-Image Retrieval.
}
\label{fig:improve_blip}
\vspace*{-2mm}
\end{figure*}

To demonstrate the effectiveness of the proposed Semantic and Visual Contrastive loss, we train BLIP ViT-B with the same training details as the original paper, except for the inclusion of SC and VC losses.
As shown in Figure~\ref{fig:improve_blip}, employing SC and VC losses significantly improves Recall@1 and RSMS for both image-to-text (I2T) and text-to-image (T2I) retrieval tasks.
Especially, a substantial improvement is observed in the Rand-voca setting (e.g., Recall@1: 48.4 $\rightarrow$ 68.2 and RSMS: 44.1 $\rightarrow$ 17.5). This improvement can be attributed to our strategy of constructing corrupted captions using random selections from the BERT vocabulary. 
In addition, interestingly, in image-to-text retrieval, using not only the SC loss, which is explicitly designed to learn semantic details, but also the VC loss, has led to further performance improvement. This observation suggests that enhancing the robustness of image embeddings can help training a robust alignment of embeddings across both modalities. 

\vspace{-2mm}
\section{Conclusion and Discussion}
\vspace{-2mm}
The paper discovers that current vision-language models are vulnerable to capturing visual and semantic details, often focusing instead on specific words or spatial regions. To stress-test VL models, we propose a new methodology for creating a robustness benchmark and demonstrate consistent degradation across various state-of-the-art models using our RoCOCO dataset. To address this issue, we introduce semantic and visual contrastive losses.
We hope that our findings provide insights for improving the robustness of VL models and for developing more diverse stress-test methods.

\textbf{Limitations.}
One limitation is that the process of randomly replacing words can result in unnatural sentences, such as ``A \textit{war} on bicycle riding next to a train (man $\rightarrow$ war).'' Additionally, the altered images are not natural and can be considered out-of-distribution data. Therefore, future work should explore methods for generating more realistic and challenging texts and images, potentially using large language models like GPT-4 or generative models such as Stable Diffusion~\cite{Rombach_2022_CVPR}.
Another limitation is that the four models used in adversarial caption generation may be more susceptible to performance degradation.
However, among the 14 models we evaluated, there was no significant difference between the 4 models used for dataset generation and the 10 models that were not involved in data generation. This could be attributed to the fact that many current models use common frozen language models like BERT. However, the trends in future models may differ, and further research should explore more generalizable methods for generating sentences.



\section*{Acknowledgments}
I would like to express my appreciation to Professor Jin Young Choi for his invaluable guidance and support.

\bibliographystyle{splncs04}
\bibliography{main}

\clearpage
\appendix
\onecolumn
\captionsetup[figure]{skip=0pt}
\setcounter{figure}{10}
\setcounter{table}{6}
\renewcommand*{\thesection}{\Alph{section}}
\renewcommand{\theequation}{\thesection.\arabic{equation}}

\begin{center}
\bf {\LARGE Appendix}
\end{center}

\section{Statistics of Source Words Selected by EI scores.}\label{appendix:sec:modelvariation}
\begin{figure*}[h]
\centering
\includegraphics[width=0.9\linewidth]{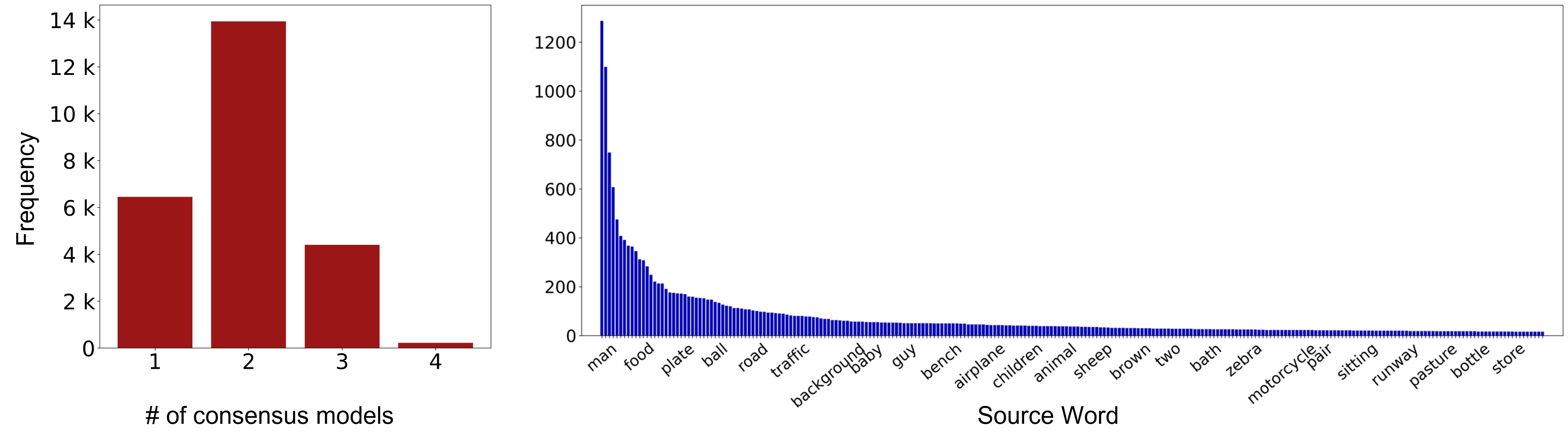}
\vspace*{-0.cm}
\hspace*{0.62cm} {\small (a) Consensus among models.} \hspace*{0.9cm} {\small (b) Source word distribution.}\hspace*{1.2cm}
\vspace*{3mm}
\caption{\textbf{Statistics of source words selected by EI scores.}
(a) Consensus among models. $x$-axis indicates the number of consensus models that pick the same word, $y$-axis the frequency of consensus for each case in $x$-axis over all captions. (b) The source word distribution exhibits a long-tailed distribution.
} 
\label{fig:ei_stats}
\end{figure*}

Figure~\ref{fig:ei_stats} represents the statistics of source words selected based on the lowest EI scores across four models.
Figure~\ref{fig:ei_stats} (a) displays the level of agreement among models in selecting the word with the lowest EI scores.
The x-axis represents the maximum number of agreements among the four models in selecting the word with the lowest EI score, while the y-axis represents the number of captions.
Interestingly, in over 70\% of cases, two or more models select the same word, despite being trained using different architectures and datasets (e.g., more pre-training data).
Furthermore, Figure~\ref{fig:ei_stats} (b) illustrates the distribution of the selected source word, which exhibits a long-tailed distribution.
This highlights a common vulnerability in the current image-text matching approach, suggesting that attacks can have a universal impact.

\section{Text-to-Image Retrieval}\label{appendix:sec:t2i}
\begin{figure*}[t]
\centering
\begin{subfigure}{0.8\linewidth}
    \includegraphics[width=\linewidth]{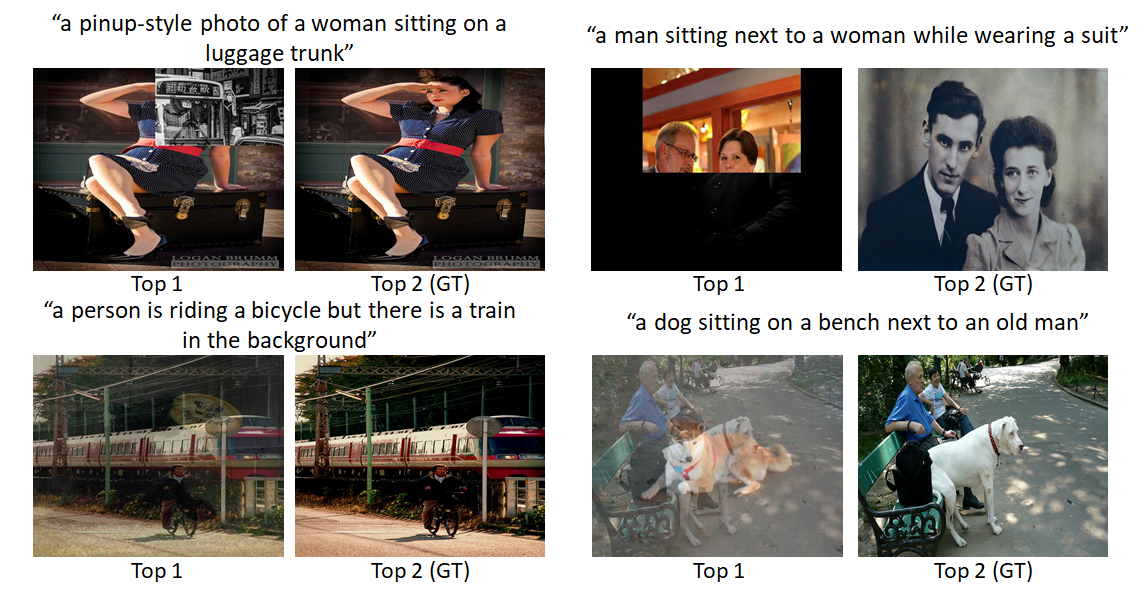}
    \caption{$\lambda = 0.6$}
    \end{subfigure}
    \begin{subfigure}{0.8\linewidth}
    \includegraphics[width=\linewidth]{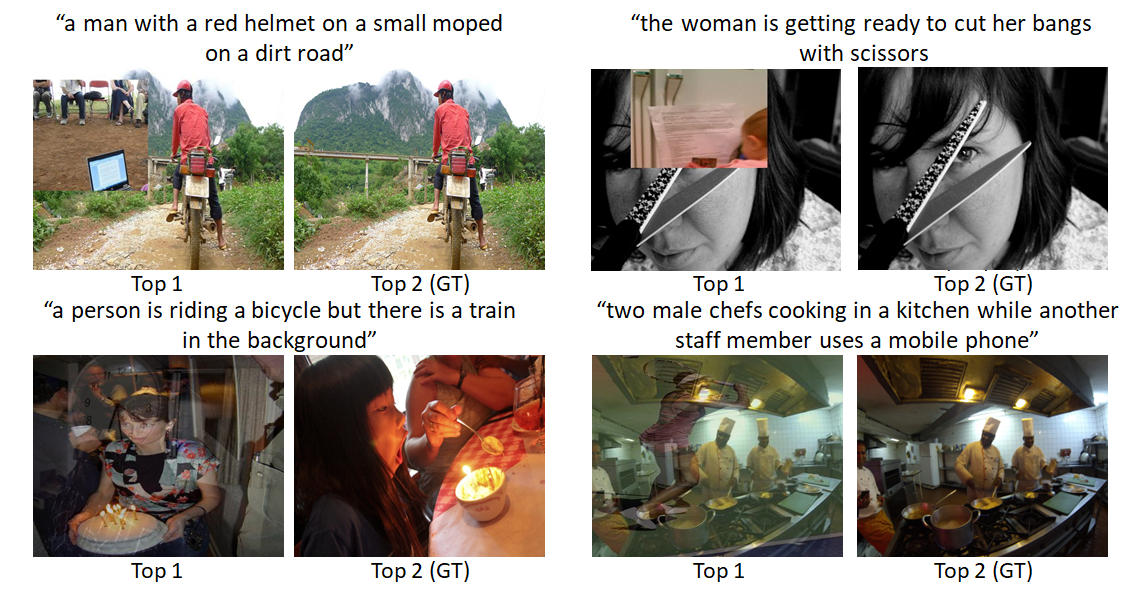}
    \caption{$\lambda = 0.7$}
    \end{subfigure}
\caption{\textbf{Text-to-Image retrieval examples.}}
\label{fig:sup_t2i_example}
\end{figure*}

We report the results on new image set with $\lambda=0.7, 0.6$ in Table~\ref{table:supp-t2i}.
We can still observe meaningful performance drop, when the added images are more significantly perturbed that seemed less confusing. 
In most cases, Incorrect Recall@1 exceeded 10\%.
In BLIP~\cite{ref:li_blip_2022}, performance degradation occurred more in fine-tuned models than in zero-shot models.
We conjecture that this is because the models overfitted to COCO dataset during finetuning.
We display the examples of retrieving incorrect images with BLIP ViT-B when $\lambda=0.6, 0.7$ in Figure~\ref{fig:sup_t2i_example}.

\begin{table*}[t]
\caption{\textbf{Text-to-Image retrieval.}
Models are evaluated with our new benchmark: Mix and Patch with different $\lambda$. 
Recall@1 (R@1)($\uparrow$), drop rate($\downarrow$), Incorrect Recall@1 (IR@1)($\downarrow$) are shown. 
The results are averaged over image generations with three different random seeds.
We can see consistent degradation across all vision-language models. 
}
\label{table:supp-t2i} 
\centering
\small
\resizebox{1\linewidth}{!}
{
\begin{tabular}{lclccclccclccclccc}
\toprule
                          & COCO 5K &  & \multicolumn{3}{c}{Mix ($\lambda=0.7$)} &  & \multicolumn{3}{c}{Mix ($\lambda=0.6$)} &  & \multicolumn{3}{c}{Patch ($\lambda=0.7$)} &  & \multicolumn{3}{c}{Patch ($\lambda=0.6$)} \\ \cline{2-2} \cline{4-6} \cline{8-10} \cline{12-14} \cline{16-18} 
                          & R@1     &  & R@1        & drop rate      & IR@1      &  & R@1        & drop rate      & IR@1      &  & R@1         & drop rate      & IR@1       &  & R@1         & drop rate      & IR@1       \\ \midrule
\multicolumn{18}{l}{\textbf{Large-scale VL pre-training models}}                                                                                                                                                                     \\
CLIP ViT-B/32 (zero-shot)~\cite{ref:clip_2021} & 30.14   &  & 25.24      & \red{-16.25\%}          & 19.16     &  & 26.87      & \red{-10.84\%}          & 14.34     &  & 25.18       & \red{-16.45\%}          & 20.27      &  & 25.96       & \red{-13.86\%}          & 17.97      \\
CLIP ViT-B/16 (zero-shot)~\cite{ref:clip_2021} & 33.03   &  & 26.60      & \red{-19.46\%}          & 22.48     &  & 28.67      & \red{-13.19\%}          & 16.90     &  & 26.14       & \red{-20.85\%}          & 25.64      &  & 27.12       & \red{-17.88\%}          & 22.76      \\
CLIP ViT-L/14 (zero-shot)~\cite{ref:clip_2021} & 36.14   &  & 30.45      & \red{-15.75\%}          & 18.59     &  & 32.01      & \red{-11.43\%}          & 15.17     &  & 30.33       & \red{-16.08\%}          & 21.16      &  & 30.96       & \red{-14.34\%}          & 19.30      \\
ALBEF~\cite{ref:albef_2021}                     & 60.67   &  & 52.53      & \red{-13.42\%}          & 14.44     &  & 55.91      & \red{-7.85\%}           & 9.19      &  & 53.71       & \red{-11.47\%}          & 12.54      &  & 54.75       & \red{-9.76\%}           & 10.92      \\
BLIP ViT-B (zero-shot)~\cite{ref:li_blip_2022}    & 56.36   &  & 48.12      & \red{-14.62\%}          & 16.52     &  & 50.81      & \red{-9.85\%}           & 11.70     &  & 46.97       & \red{-16.66\%}          & 18.74      &  & 48.38       & \red{-14.16\%}          & 16.33      \\
BLIP ViT-B~\cite{ref:li_blip_2022}                & 64.31   &  & 53.56      & \red{-16.72\%}          & 20.15     &  & 57.77      & \red{-10.18\%}          & 13.45     &  & 55.20       & \red{-14.17\%}          & 16.79      &  & 56.68       & \red{-11.87\%}          & 14.82      \\
BLIP ViT-L (zero-shot)~\cite{ref:li_blip_2022}    & 58.18   &  & 51.21      & \red{-11.98\%}          & 13.82     &  & 53.69      & \red{-7.71\%}           & 9.38      &  & 51.05       & \red{-12.25\%}          & 13.96      &  & 52.28       & \red{-10.14\%}          & 11.95      \\
BLIP ViT-L~\cite{ref:li_blip_2022}                & 65.06   &  & 52.06      & \red{-19.98\%}          & 24.43     &  & 57.08      & \red{-12.27\%}          & 16.19     &  & 55.58       & \red{-14.57\%}          & 18.05      &  & 57.19       & \red{-12.10\%}          & 15.41      \\ \midrule
\multicolumn{18}{l}{\textbf{Visual Semantic Embedding models}}                                                                                                                                                                       \\
VSRN~\cite{ref:li_vsrn_2019}                      & 40.34   &  & 34.80      & \red{-13.72\%}          & 19.24     &  & 37.04      & \red{-8.17\%}           & 12.69     &  & 34.01       & \red{-15.68\%}          & 21.31      &  & 34.99       & \red{-13.25\%}          & 18.59      \\
SAF~\cite{ref:sgraf_2021}                       & 40.11   &  & 35.55      & \red{-11.37\%}          & 16.57     &  & 36.91      & \red{-7.98\%}           & 12.32     &  & 35.22       & \red{-12.20\%}          & 16.81      &  & 35.75       & \red{-10.87\%}          & 15.24      \\
SGR~\cite{ref:sgraf_2021}                       & 40.45   &  & 35.59      & \red{-12.01\%}          & 16.54     &  & 37.23      & \red{-7.96\%}           & 11.96     &  & 35.23       & \red{-12.90\%}          & 17.12      &  & 35.85       & \red{-11.37\%}          & 15.53      \\
VSE (BUTD region)~\cite{ref:chen_vse_infty_2021}         & 42.46   &  & 38.74      & \red{-8.76\%}           & 11.99     &  & 40.38      & \red{-4.90\%}           & 7.20      &  & 38.18       & \red{-10.08\%}          & 13.57      &  & 38.99       & \red{-8.17\%}           & 11.42      \\
VSE (BUTD grid)~\cite{ref:chen_vse_infty_2021}           & 44.07   &  & 39.01      & \red{-11.47\%}          & 15.39     &  & 41.22      & \red{-6.46\%}           & 9.26      &  & 40.10       & \red{-9.00\%}           & 12.12      &  & 40.94       & \red{-7.09\%}           & 9.94       \\
VSE (WSL grid)~\cite{ref:chen_vse_infty_2021}            & 51.55   &  & 45.13      & \red{-12.46\%}          & 15.70     &  & 47.97      & \red{-6.95\%}           & 9.30      &  & 48.37       & \red{-6.17\%}           & 8.02       &  & 48.93       & \red{-5.08\%}           & 6.58       \\ \bottomrule
\end{tabular}
}
\end{table*}

\section{Substituting more words}\label{appendix:sec:morewords}
The results in Table~\red{6} show meaningful performance degradation across the entire model, even when the original semantic meaning is significantly disrupted.
When more words are substituted to create a stranger sentence (e.g., 5 words substitution), a large-scale vision-language pre-training models demonstrate relative robustness compared to the visual semantic embedding models.



\begin{figure*}[t]
\centering
    \begin{subfigure}{1\linewidth}
    \includegraphics[width=1\linewidth]{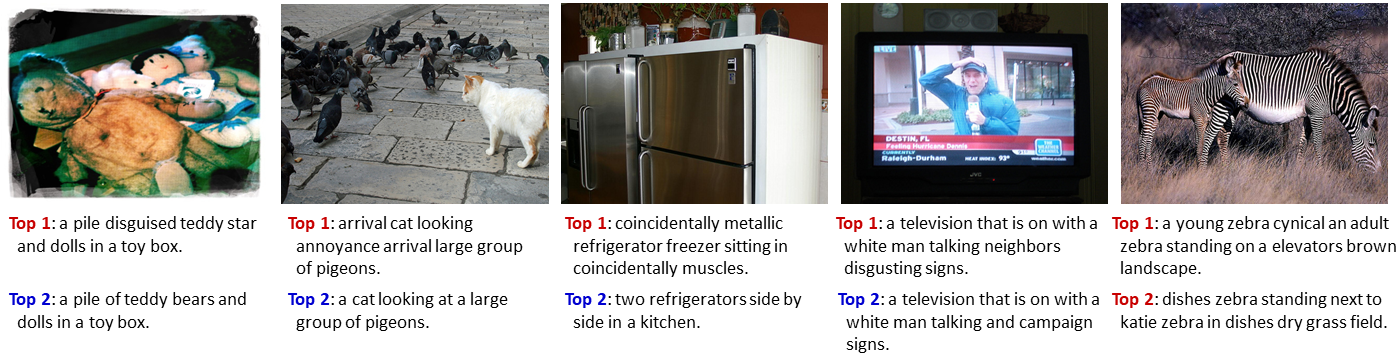}
    \caption{Two words substitution}
    \end{subfigure}
    \begin{subfigure}{1\linewidth}
    \includegraphics[width=1\linewidth]{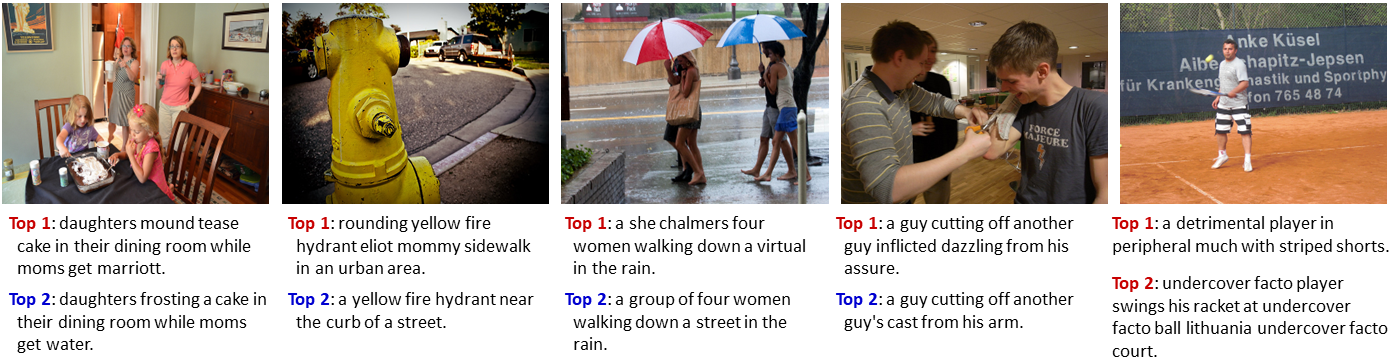}
    \caption{Three words substitution}
    \end{subfigure}
    \begin{subfigure}{1\linewidth}
    \includegraphics[width=1\linewidth]{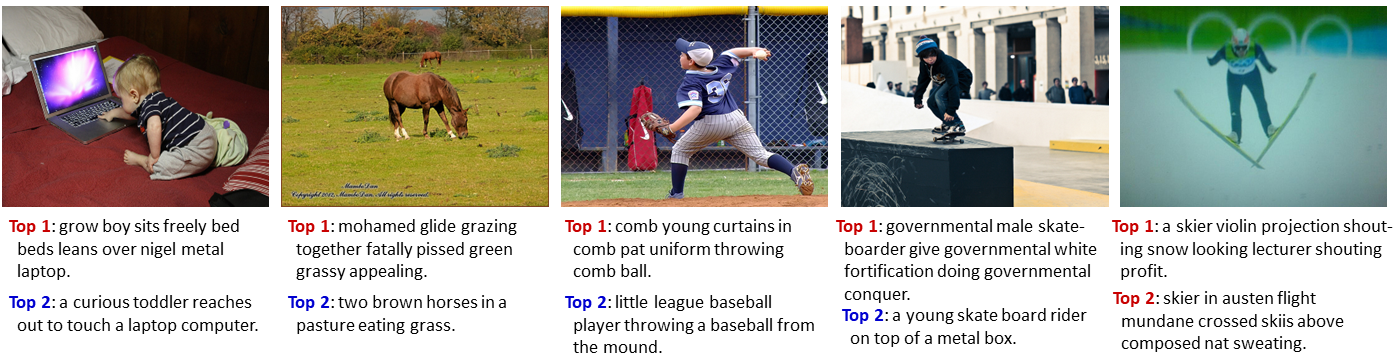}
    \caption{Five words substitution}
    \end{subfigure}
\caption{\textbf{Examples of substituting multiple random words with BLIP (ViT-B).}
   }
\label{fig:sup_multi_ex}
\end{figure*}

Figure~\ref{fig:sup_multi_ex} shows examples of newly added captions that BLIP ViT-B model has retrieved as top 1. 
While the created captions are not natural, they include some keywords.
Thus, we can conclude that the model is focusing on some nouns rather than the whole sentence.

\end{document}